\documentclass[10pt,twocolumn,letterpaper]{article}

\usepackage{iccv}
\usepackage{times}
\usepackage{epsfig}
\usepackage{graphicx}
\usepackage{amsmath}
\usepackage{amssymb}
\usepackage{bbm}
\usepackage{subcaption}
\usepackage{multirow}
\usepackage{adjustbox}
\usepackage[dvipsnames]{xcolor}
\usepackage{tabu}
\usepackage{floatrow}
\usepackage{booktabs}
\usepackage{makecell}
\usepackage{epigraph}
\usepackage{mathtools}

\usepackage{algorithm}
\usepackage{algorithmic}
\usepackage{tabularx}
\usepackage{gensymb}

\usepackage{animate}

\usepackage[percent]{overpic}

\usepackage[pagebackref=true,breaklinks=true,letterpaper=true,colorlinks,bookmarks=false]{hyperref}

\definecolor{darkred}{rgb}{0.7,0.1,0.1}
\definecolor{darkgreen}{rgb}{0.1,0.7,0.1}
\definecolor{cyan}{rgb}{0.7,0.0,0.7}
\definecolor{dblue}{rgb}{0.2,0.2,0.8}
\definecolor{maroon}{rgb}{0.76,.13,.28}
\definecolor{burntorange}{rgb}{0.81,.33,0}

\ifdefined\ShowNotes
  \newcommand{\colornote}[3]{{\color{#1}\bf{#2: #3}\normalfont}}
\else
  \newcommand{\colornote}[3]{}
\fi

\iccvfinalcopy 

\def\iccvPaperID{8809} 

\ificcvfinal\fi
\begin{document}

\title{GAN-Control: Explicitly Controllable GANs}

\author{
    Alon Shoshan\qquad Nadav Bhonker\qquad Igor Kviatkovsky\qquad G\'{e}rard Medioni\\
    Amazon \\
    \tt\small \{alonshos, nadavb, kviat, medioni\}@amazon.com \\
}


\twocolumn[{%
\renewcommand\twocolumn[1][]{#1}%
\maketitle
\ificcvfinal\thispagestyle{empty}\fi
\begin{center}
\centering
\begin{tabular}{c c c}
    \includegraphics[height=0.085\linewidth]{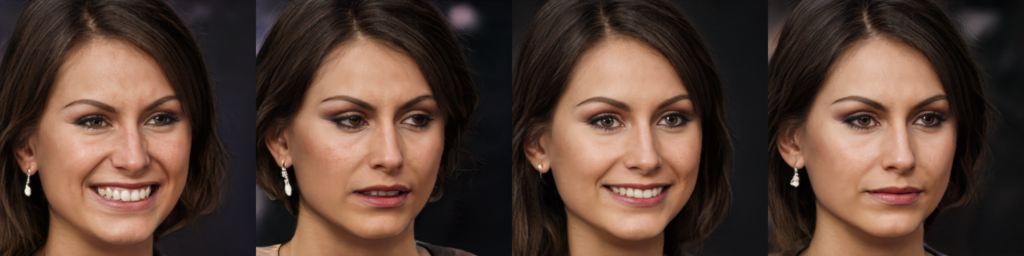} &
    \includegraphics[height=0.085\linewidth]{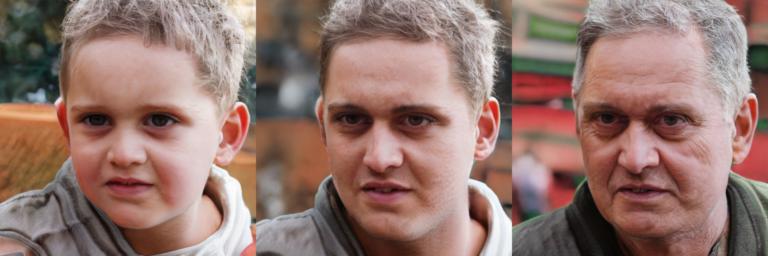} &
    \includegraphics[height=0.085\linewidth]{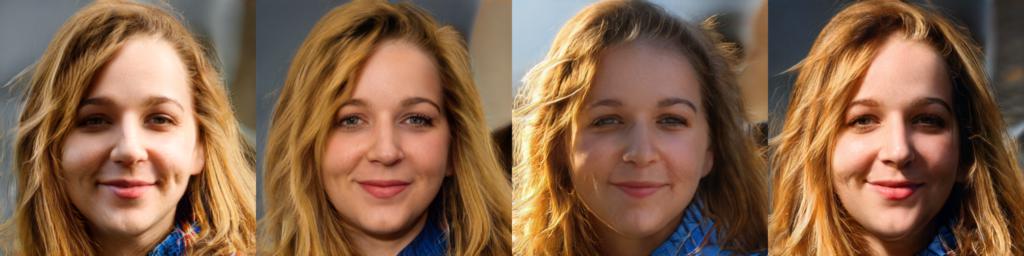} \\

\end{tabular}
\begin{tabular}{c c c}
    \includegraphics[height=0.085\linewidth]{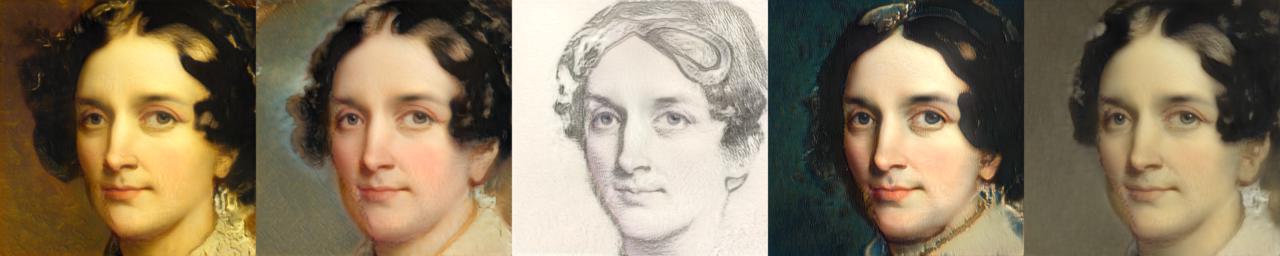} & 
    \includegraphics[height=0.085\linewidth]{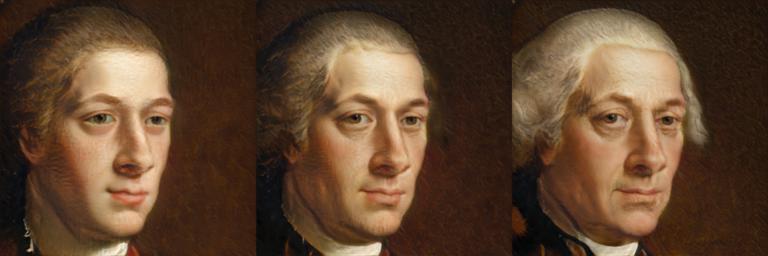} &
    \includegraphics[height=0.085\linewidth]{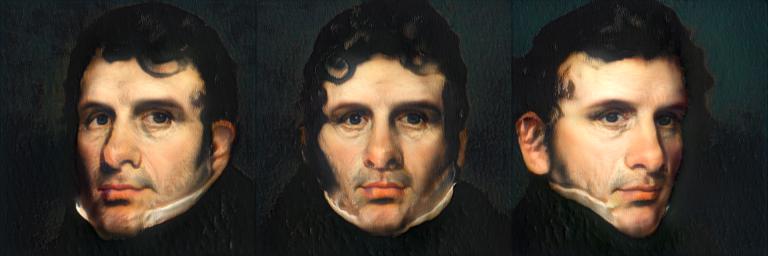} \\

\end{tabular}
\begin{tabular}{c c c}
    \includegraphics[height=0.085\linewidth]{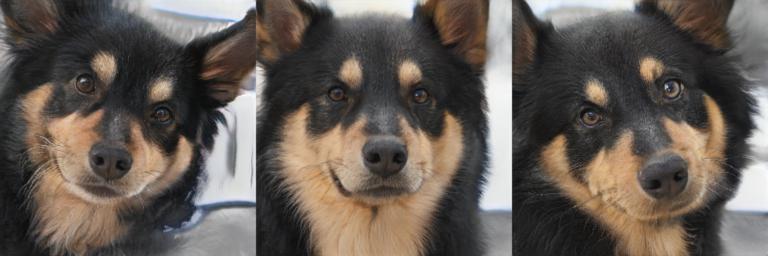} &
    \includegraphics[height=0.085\linewidth]{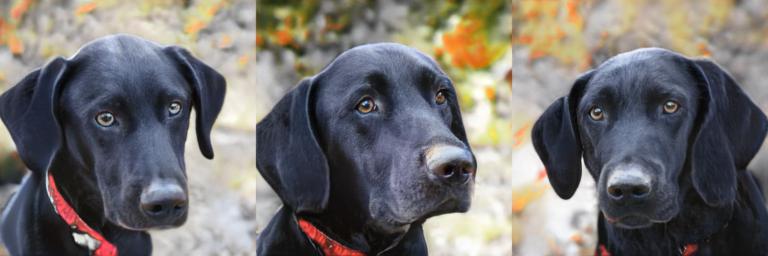} &
    \includegraphics[height=0.085\linewidth]{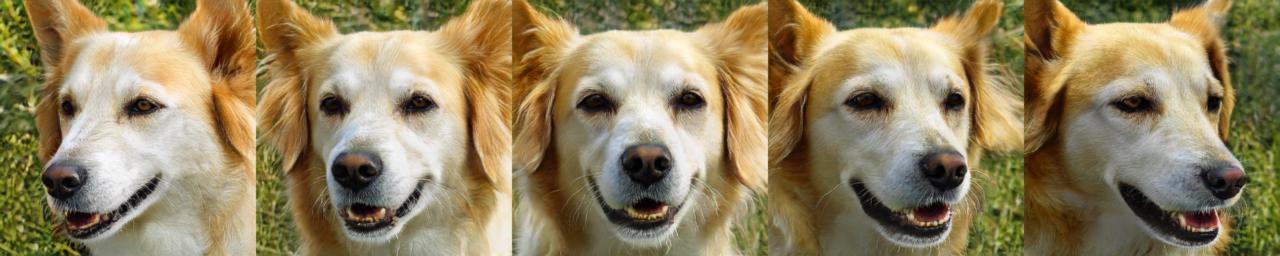} \\ 
\end{tabular}
\captionof{figure}{We propose a framework for training GANs in a disentangled manner which allows for explicit control over generation attributes.
Our method is applicable to diverse controls in various domains. 
First row (left to right) demonstrates our control over facial expression, age and illumination of human portraits.
Second row (left to right) demonstrates our control over artistic style, age and pose of paintings. Third row demonstrates our pose control over faces of dogs. 
}
\label{fig:teaser}
\end{center}
}]

\begin{abstract}
    We present a framework for training GANs with explicit control over generated facial images.
    We are able to control the generated image by settings exact attributes such as age, pose, expression, etc.
    Most approaches for manipulating GAN-generated images achieve partial control by leveraging the latent space disentanglement properties, obtained implicitly after standard GAN training. 
    Such methods are able to change the relative intensity of certain attributes, but not explicitly set their values.
    Recently proposed methods, designed for explicit control over human faces, harness morphable 3D face models (3DMM) to allow fine-grained control capabilities in GANs.
    Unlike these methods, our control is not constrained to 3DMM parameters and is extendable beyond the domain of human faces.
    Using contrastive learning, we obtain GANs with an explicitly disentangled latent space. 
    This disentanglement is utilized to train control-encoders mapping human-interpretable inputs to suitable latent vectors, thus allowing explicit control.
    In the domain of human faces we demonstrate control over identity, age, pose, expression, hair color and illumination.
    We also demonstrate control capabilities of our framework in the domains of painted portraits and dog image generation.
    We demonstrate that our approach achieves state-of-the-art performance both qualitatively and quantitatively.
\end{abstract}


\section{Introduction}
\label{sec:introduction}
Generating controllable photorealistic images has applications spanning a variety of fields such as cinematography, graphic design, 
video games, medical imaging, virtual communication and ML research. 
For faces in particular, impressive breakthroughs were made. 
As an example, in the film industry, computer generated characters are replacing live actor footage. 
Earlier work on controlled face generation primarily relied on 3D face rig modeling~\cite{li2017learning, sanyal2019learning}, controlled by 3D morphable face model parameters, such as 3DMM~\cite{blanz1999morphable, egger20203d}. 
While easily controllable, such methods tend to suffer from low photorealism. 
Other methods that rely on 3D face scanning techniques may provide highly photorealistic images, but at a significant cost and limited variability.
Recent works on high resolution images synthesis using generative adversarial networks (GANs)~\cite{goodfellow2014generative} have demonstrated the ability to generate photorealistic faces of novel identities, indistinguishable from those of real humans~\cite{karras2017progressive,karras2019style,Karras2019stylegan2}.
However, these methods alone lack interpretability and control over the generative process, compared to the 3D graphic alternatives.

These results have inspired the community to explore ways to benefit from both worlds -- generating highly photorealistic faces using GANs while controlling their fine-grained attributes, such as pose, illumination and expression with 3DMM-like parameters.
Deng~\etal~\cite{deng2020disentangled}, Kowalski~\etal~\cite{KowalskiECCV2020} and Tewari~\etal~\cite{Tewari_2020_CVPR} introduce explicit control over GAN-generated faces, relying on guidance from 3D face generation pipelines. 
Along with the clear benefits, such as the precise control and perfect ground truth, reliance on such 3D face models introduces new challenges. 
For example, the need to overcome the synthetic-to-real domain gap~\cite{KowalskiECCV2020,deng2020disentangled}. 
Finally, all these methods' expressive power is bounded by the capabilities of the model they rely on. 
In particular, it is not possible to control human age if the 3D modeling framework does not support it. 
It is also impossible to apply the same framework to different but similar domains, such as paintings or animal faces, if these assets are not supported by the modeling framework.
All of these stand in the way of creating a simple, generic and extendable solution for explicitly controllable GANs.

In this work we present a unified approach for training a GAN to generate high-quality, controllable images. 
Specifically, we demonstrate our approach in the domains of facial portrait photos, painted portraits and dogs (see Fig.~\ref{fig:teaser}).
We depart from the use of the highly detailed 3D face models~\cite{deng2020disentangled,KowalskiECCV2020,Tewari_2020_CVPR} in favor of supervision signals provided by a set of pre-trained models, each controlling a different feature. 
We show that our approach significantly simplifies the generation framework, does not compromise image quality or control accuracy, and allows us to control additional aspects of facial appearances, which cannot be modeled by graphical pipelines.
We achieve this by combining several concepts. 
We construct the GAN's latent space as a composition of sub-spaces, each corresponding to a specific property.
During training, we enforce images generated by identical latent sub-vectors to have similar properties, as predicted by some off-the-shelf model.
Respectively, images generated by different latent sub-vectors are enforced to have different predicted properties.
As a result, disentanglement between the latent sub-spaces is achieved.
Finally, to allow for human-interpretable control, for each attribute we train an encoder converting values from its feasible range to its corresponding sub-latent space. 
As an additional application, we present a novel image projection approach suitable for disentangled latent spaces.

We summarize our contributions as following:
\begin{enumerate}
    \item We present a novel state-of-the art approach for training explicitly controllable, high-resolution GANs.
    \item Our approach is extendable to attributes beyond those supported by 3D modeling and rendering frameworks, making it applicable to additional domains.
    \item We present a disentangled projection method that enables real image editing.
\end{enumerate}

\begin{figure*}[t]

\includegraphics[width=\linewidth]{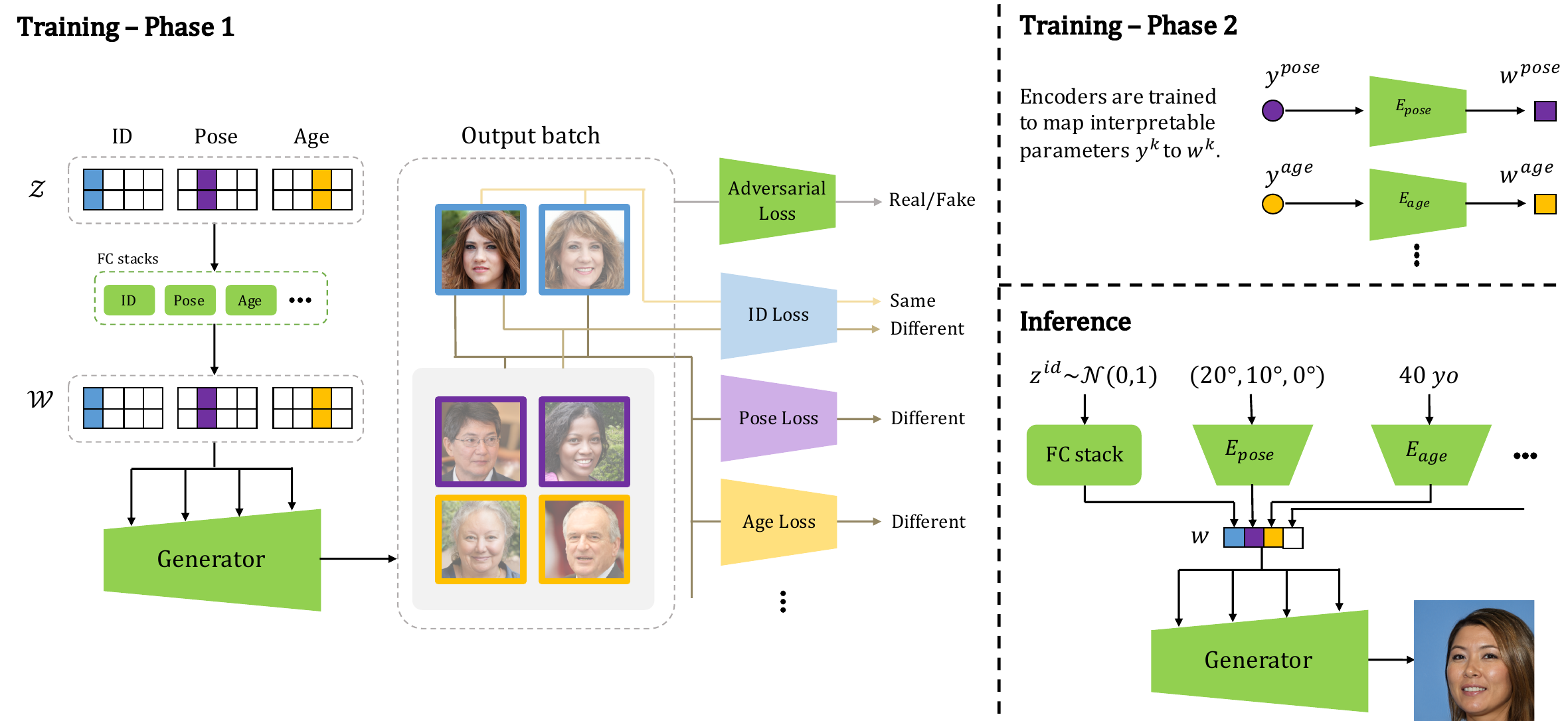}
\caption{\textbf{Explicitly controllable GAN}: 
In Phase 1, we construct every batch so that for each attribute, there is a pair of latent vectors sharing a corresponding sub-vector, $\mathbf{z}^k$.
In addition to the adversarial loss, each image in the batch is compared in a contrastive manner, attribute-by-attribute, to all others, taking into account if it has the same or a different sub-vector. 
In Phase 2, encoders are trained to map interpretable parameters to suitable latent vectors. 
Inference: An explicit control over the attribute $k$ is achieved by setting the $E_k$'s input to a required value.}
\vspace{-0.1cm}
\label{fig:pipeline}
\end{figure*}

\section{Related work}
\label{sec:related-work}
Generative adversarial networks~\cite{goodfellow2014generative} introduced new possibilities to the field of image generation and synthesis. 
Currently, state-of-the-art GANs~\cite{brock2018large,karras2017progressive,karras2019style,Karras2019stylegan2} can produce high-resolution images that are indistinguishable from real ones. 
Next, we provide an overview of different approaches to control the generated output of GANs.

\textbf{Relative control over image generation:}
A widely studied approach for controlling the generated images of GANs is by exploiting the inherent disentanglement properties of their latent space~\cite{gansteerability, yang2019semantic, harkonen2020ganspace, Shen_2020_CVPR, balakrishnan2020towards}.
H{\"a}rk{\"o}nen~\etal~\cite{harkonen2020ganspace} use principal component analysis (PCA) in latent space to identify directions that correspond to image attributes.
Shen~\etal~\cite{Shen_2020_CVPR} use off-the-shelf binary classifiers to find separation boundaries in the latent space where each side of the boundary corresponds to an opposite semantic attribute (\eg, young vs. old).
Traversing a latent vector closer to or further from a boundary translates to increasing or decreasing the corresponding attribute intensity.
While simple, these methods may exhibit entanglement, \ie, changing one attribute affects others. 
In~\cite{donahue2018sdgan, DBLP:conf/cvpr/Shen0YWT18} the above is mitigated by disentangling the GAN's latent space during training.
While the above methods allow for relative control over the generation (\eg, turn the face older or rotate the face towards the left), they do not provide explicit control (\eg, generate a $40$ years old face, rotated $30\degree$ to the left).

\textbf{Explicit control over image generation:}
Conditional GANs~\cite{cgan,Odena2017ConditionalIS,Miyato2018cGANsWP,brock2018large} have been widely employed to control the generation by incorporating a class label inference loss term.
All these works support conditioning on a single discrete (categorical) variable and are not suitable for continues variables, as was broadly discussed in Ding~\etal~\cite{ding2021ccgan}.
Furthermore, none of the above works address the problem of controlling multiple attributes at once.
Recently, three novel methods were proposed to allow fine-grained explicit control over \emph{de novo} face image generation: StyleRig~\cite{Tewari_2020_CVPR}, DiscoFaceGAN~\cite{deng2020disentangled} (DFG), and CONFIG~\cite{KowalskiECCV2020}.
These methods propose solutions for translating controls of 3D face rendering models to GAN-generating processes. 
Both StyleRig and DFG utilize 3DMM~\cite{blanz1999morphable} parameters as controls in the generation framework. 
This restricts both approaches to provide controls only over the expression, pose and illumination, while preserving identity (ID).
CONFIG uses a custom 3D image rendering pipeline to generate an annotated synthetic dataset.
This dataset is later used to acquire controls matching the synthetic ground truth,  
allowing CONFIG to add controls such as hair style and gaze.
Producing such datasets is hard and requires professional handcrafted 3D assets.
We emphasize that these methods are only applicable in the domain of human faces, and only to the controls parametrized by 3D face models. 
In contrast to the above methods, our approach does not rely on 3D face rendering frameworks.
Rather, it relies on our ability to estimate such properties.


\textbf{Image editing:}
Rather than generating images \emph{de novo}, these methods receive an image as input and manipulate its attributes either by using image-to-image translation techniques~\cite{CycleGAN2017, wang2018pix2pixHD, park2019SPADE,nizan2020council, choi2018stargan, choi2020starganv2, huang2018munit}, by incorporating pre-trained models to supervise GAN's training~\cite{FaceFeat-GAN,Bao0WLH18,AttGan,yao2020high}, 
or by projecting the image to the GAN's latent space and manipulating it~\cite{zhu2016generative,abdal2020styleflow,tewari2020pie,pan2021do,spingarn2021gan,zhuang2021enjoy,zhu2020indomain}.
Our work focuses on controllable \emph{de novo} image generation, but also allows editing real images via projection to latent space.

\section{Proposed approach}
\label{sec:proposed-approach}
In this section we present our framework for training explicitly controllable GANs.
Our approach is simple yet effective and is comprised of two phases (see Fig.~\ref{fig:pipeline}):
\begin{itemize}
    \item \textbf{Disentanglement by contrastive learning:} training a GAN with explicitly disentangled properties. 
    As a result, the latent space is divided into sub-spaces, each encoding a different image property. 
    \item \textbf{Interpretable explicit control:} for each property, an MLP encoder is trained to map control parameter values to a corresponding latent sub-space. 
    This enables explicit control over each one of the properties. 
\end{itemize}

\subsection{Disentanglement by contrastive learning} 
\label{sec:loss}
The approach builds on the StyleGAN2~\cite{Karras2019stylegan2} architecture.
Initially, we divide both latent spaces, $\mathcal{Z}$ and $\mathcal{W}$ to $N\!+\!1$ separate sub-spaces, $\{\mathcal{Z}^k\}_{k=1}^{N+1}$, and $\{\mathcal{W}^k\}_{k=1}^{N+1}$, where $N$ is the number of control properties.
Each sub-space is associated with an attribute (\eg, ID, age etc.) except for the last one.
Similarly to Deng~\etal~\cite{deng2020disentangled} the last sub-space encodes the rest of the image properties that are not controllable. 
We modify the StyleGAN2 architecture so that each control has its own 8-layered MLP.
We denote $\mathbf{z}=(\mathbf{z}^1 \mathbf{z}^2 \ldots \mathbf{z}^{N+1})$ and $\mathbf{w}=(\mathbf{w}^1 \mathbf{w}^2 \ldots \mathbf{w}^{N+1})$ the concatenation of the sub-vectors in both latent spaces.
The combined latent vector, $\mathbf{w}$, is then fed into the generator.

Next, we describe how we enforce disentanglement during training.
Let $\mathbf{\cal I}_i= G(\mathbf{z}_i)$ denote an image generated from a latent vector $\mathbf{z}_i$ and let $B = \{ \mathbf{z}_i \}_{i=1}^{N_B} $ denote a latent vector batch of size $N_B$. 
We define our factorized-contrastive loss as:
\begin{equation}
L_c = \sum_{\substack{\mathbf{z}_i,\mathbf{z}_j \in B\\ i \ne j}}\sum_{k=1}^N l_k(\mathbf{z}_i,\mathbf{z}_j),    
\end{equation}
where $l_k$ is a contrastive loss component for attribute $k$. We define the per-attribute contrastive loss as,  
\begin{equation}
\small
l_k(\mathbf{z}_i,\mathbf{z}_j)\!=\! 
\begin{cases}
    \frac{1}{C_k^{+}} \max{(d_k(\mathbf{\cal I}_i,\mathbf{\cal I}_j) - \tau_k^+, 0)},&  \mathbf{z}_i^k\!=\!\mathbf{z}_j^k \\
    \frac{1}{C_k^\texttt{-}} \max{(\tau_k^- - d_k(\mathbf{\cal I}_i,\mathbf{\cal I}_j), 0)},& \text{otherwise}
\end{cases}
\end{equation}
where $\mathbf{z}_i^k$ denotes the $k$-th sub-vector of $\mathbf{z}_i$, $d_k$ is the distance function for attribute $k$, $\tau_k^{\pm}$ are the per-attribute thresholds associated with same and different sub-vectors and $C_k^{\pm}$ are constants that normalize the loss according to the number of same and different loss components, 
\ie $C_k^{+} = \sum_{i,j} {\mathbbm{1}\{\mathbf{z}_i^k = \mathbf{z}_j^k\}}$ and  $C_k^\texttt{-} = \sum_{i,j} {\mathbbm{1}\{\mathbf{z}_i^k \neq \mathbf{z}_j^k\}}$.

We construct each training batch to contain pairs of latent vectors that share one sub-vector, \ie, for each attribute, $k \in \{1,\dots,N\}$, we create a pair of latent vectors, $\mathbf{z}_i$ and $\mathbf{z}_j$,  where $\mathbf{z}_i^k=\mathbf{z}_j^k$ and $\mathbf{z}_i^r \ne \mathbf{z}_j^r$ for $r \in \{1,\ldots,N+1\}, r \ne k$.
For example, let us assume that the generator has produced a batch of size $N_b\!>\!2$, where images ${\cal I}_0$ and ${\cal I}_1$ share the same $\mathbf{z}^{ID}$ (see the pair of images with the blue frame in Fig.~\ref{fig:pipeline}).
The ID component of the contrastive loss, $l_{ID}$, will penalizes the dissimilarities between the ${\cal I}_0$'s and ${\cal I}_1$'s IDs and the similarities between ${\cal I}_0$'s or ${\cal I}_1$'s ID to the IDs of all other images in the batch.
The other loss components (age, pose, \etc) will penalize for similarity between ${\cal I}_0$ and any other image in the batch.
The losses for all other images in the batch are constructed in the same manner.

To be able to control a specific attribute of the generated image, we assume that we are given access to a differentiable function $M_k: \mathcal{I} \rightarrow \mathbb{R}^{D_k}$, mapping an image to a $D_k$-dimensional space. 
We assume that the projected images with similar attribute values fall close to each other, and images with different attribute values fall far from one another.
Such requirements are met by most neural networks trained with either a classification or a regression loss -- for example, a model estimating the head pose or the person's age.  
We define the $k$'s attribute distance between two images ${\cal I}_i$ and ${\cal I}_j$ as their distance in the corresponding embedding space: 
\begin{align}
    d_{k}({\cal I}_i, {\cal I}_j) = dist(M_k({\cal I}_i), M_k({\cal I}_j)),
\end{align}
where $dist(\cdot,\cdot)$ is a distance metric, \eg, $L_1$, $L_2$, cosine-distance, \etc.
For example, to capture the ID property, a face recognition model, $M_{ID}$, is used to extract embedding vectors from the generated images. 
Then, the distances between the embedding vectors are computed using the cosine-distance.

In Section~\ref{sec:experiments} we demonstrate that as the result of training with this architecture and batch sampling protocol, we achieve disentanglement in the GAN's latent space.
While such disentanglement allows to assign a randomly sampled value to each individual attribute, independently of the others, additional work is required for turning such control explicit and human-interpretable, \eg, generate a human face image with a specific user-defined age.

\subsection{Interpretable explicit control}
\label{sec:explicit_control}
We propose a simple procedure to allow explicit control of specific attributes. 
We train a mapping $E_k: y^k \rightarrow \mathbf{w}^k$, where $y^k$ is a human-interpretable representation of the attribute (\eg, age $= 20$yo, pose $= (20\degree, 5\degree, 2\degree)$, \etc). 
Given a trained disentangled GAN, we train $N$ encoders $\{E_k\}_{k=1}^N$, one for each attribute (see Training-Phase 2 in Fig.~\ref{fig:pipeline}).
Then, at inference time we can synthesize images using any combination of sub-vectors $ \{\mathbf{w}^k\}_{k=1}^{N+1} $, where $\mathbf{w}^k$ is either controlled explicitly using $E_k$ or sampled from $\mathbf{z}^k$ and consequently mapped to $\mathbf{w}^k$ (see Inference in Fig.~\ref{fig:pipeline}).

To train the control encoders, we randomly sample $N_s$ latent vectors $\{\mathbf{z}_i\}_{i=1}^{N_s}$ and map them to the intermediate latent vectors, $\{\mathbf{w}_i\}_{i=1}^{N_s}$.
Then, for each attribute, $k$, we map $\mathbf{z}_i$ to a predicted attribute value $y^{k}_i=Q_k(M_k(G(\mathbf{z}_i)))$, where $Q_k(M_k(\cdot))$ is equivalent to applying the attribute predictor.
Thus we obtain $N$ distinct datasets $\{\{\mathbf{w}_i^k, y_i^k\}_{i=1}^{N_s}\}_{k=1}^N$,
where for each intermediate sub-vector $\mathbf{w}^k$ there is a corresponding attribute predicted from the image it produced. We then train $N$ encoders, each on its corresponding dataset. 
In our experiments we show that despite its simplicity, our encoding scheme does not compromise control accuracy compared to other methods.
 
  
\section{Experiments}
\label{sec:experiments}
In this section we present experiments on the domain of faces and paintings that demonstrate the flexibility of the proposed approach.
Additional experiments for images of dogs are presented in the supplementary.
We quantitatively compare our approach to recent published approaches.

\subsection{Face generation}
\textbf{Implementation details:} We use the FFHQ dataset~\cite{karras2019style} downsampled to 512x512 resolution.
The latent spaces $ {\cal Z} $ and $ {\cal W} $ are divided into the following sub-spaces: ID, pose, expression, age, illumination, hair color and ``other''.
Next, we list the models, $M_k$, that were used to compute the distance measures, $d_k$, for each one of the attributes. 
For the ID, head-pose, expression, illumination, age and hair color we used ArcFace~\cite{deng2019arcface}, Ruiz~\etal~\cite{Ruiz_2018_CVPR_Workshops}, ESR~\cite{SMW20}, the $\gamma$ output of R-Net~\cite{deng2019accurate}, Dex~\cite{Rothe-ICCVW-2015}, average color of hair segmented by PSPNet~\cite{zhao2017pspnet}, respectively (additional details in supplementary material). 
In the second phase (Section~\ref{sec:explicit_control}), we train five encoders ($E_{pose}, E_{exp}, E_{age}, E_{illum}, E_{hair}$), each composed of a 4-layered MLP.
The input to our control encoder is defined as follows: $y^{age}\!\in\![15,75]$ years-old (yo), 
$y^{pose}\!\in\!\left[-90\degree,90\degree\right]^3$ is represented by the Euler angles $\theta= \{\text{Pitch}, \text{Yaw}, \text{Roll}\}$, 
$y^{illum}\!\in\!{\rm I\!R}^{27}$ is represented by the $\gamma$ Spherical Harmonics (SH) coefficients approximating scene illumination~\cite{Ramamoorthi2001AnER}, $y^{exp}\!\in\!{\rm I\!R}^{64}$ is represented by the $\beta$ expression coefficients of the 3DMM~\cite{blanz1999morphable} model, $y^{hair} \!\in\! [0,255]^{3}$ is represented by the mean RGB values.
\begin{figure}[]

\includegraphics[width=0.98\linewidth]{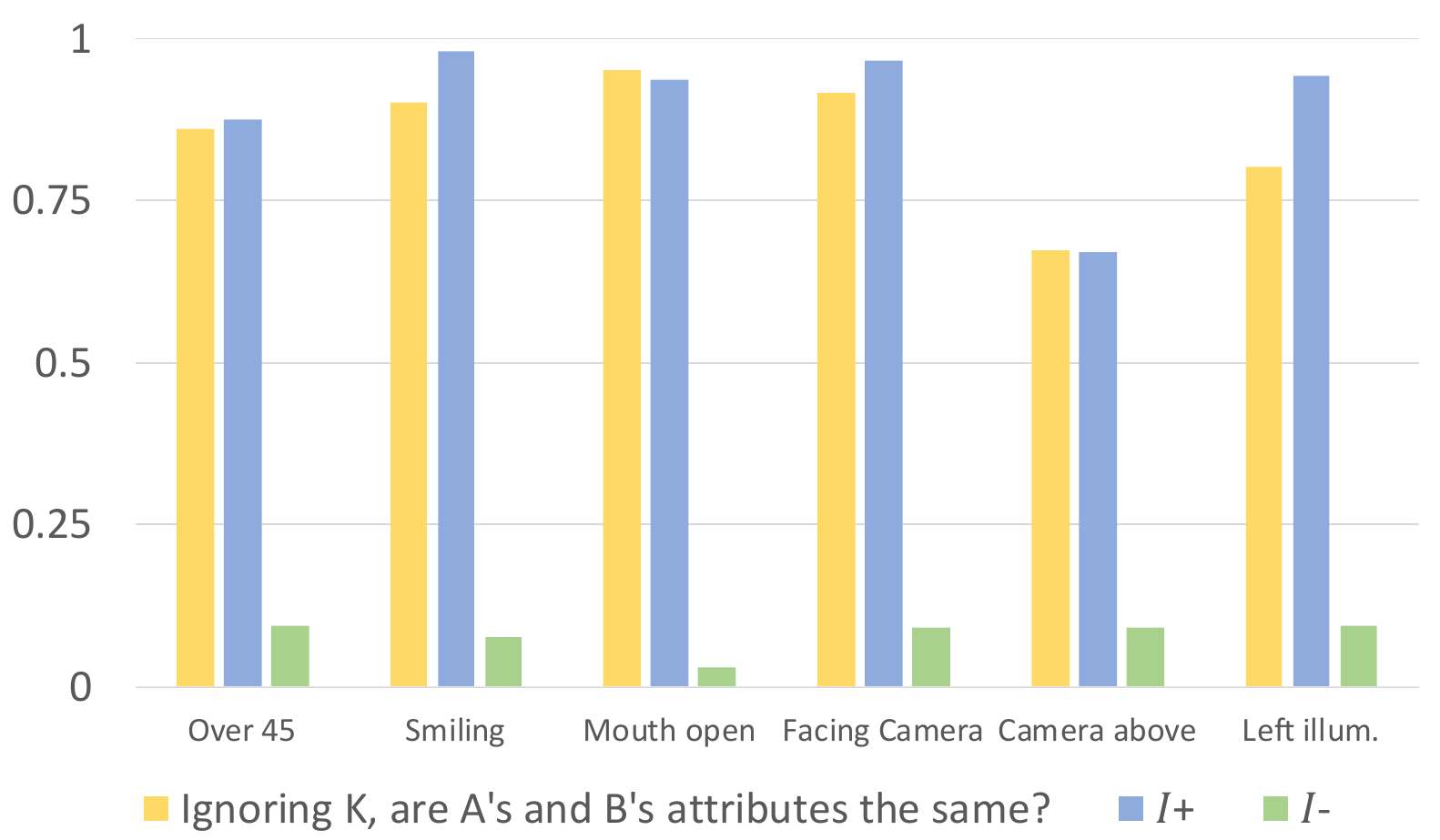}
\vspace{-0.2cm}
\caption{\textbf{Disentanglement user study:} 
Blue and green bars show whether users agree that a given attribute is present or lacking in images (${\cal I}_{+}$, $\uparrow$), (${\cal I}_\texttt{-}$, $\downarrow$) respectively.
The yellow bars measure whether the users agree that the other attributes are maintained.}
\label{fig:user_study}
\end{figure}
\begin{table}
    \setlength{\textwidth}{0pt}
    \newcolumntype{?}{!{\vrule width 1pt}}
    \centering
    \small
	\begin{tabular}{l?c|c|c}
    \toprule
	\textbf{GAN} & Ours & DFG~\cite{deng2020disentangled} & CONFIG~\cite{KowalskiECCV2020} \\
	 \textbf{Version} & 512x512 &  256x256 & 256x256\\ 
    \hline
    Vanilla & 3.32 & 5.49 & 33.41\\
    Controlled & 5.72 & 12.9 & 39.76\\
    \bottomrule
    \end{tabular}
    \caption{\textbf{FID$\downarrow$ score for different methods on FFHQ:} second row shows the dataset resolution. Note that the FID scores cannot be compared between columns since every method uses different pre-processing for the FFHQ dataset (\eg, image size, alignment, cropping).  
    }
    \label{tab:fid}
\end{table}
\begin{table}
    \setlength{\textwidth}{0pt}
    \newcolumntype{?}{!{\vrule width 1pt}}
    \centering
    \small

	\begin{tabular}{l?c|c|c}
        \toprule
	    & Ours & DFG & CONFIG
	    \\
        \hline
        Synthetic comparison& $\pmb{67\%}$ & $22\%$ & $11\%$
        \\
        Synthetic vs. real& $\pmb{47\%}$ & $27\%$ & $16\%$
        \\
        \bottomrule
    \end{tabular}
    \caption{\textbf{Photorealism user studies$\uparrow$:} (First row) users were asked to vote for the most realistic image from triplets of synthetic images (Ours, DFG, CONFIG).
    (Second row) users were shown pairs of images -- one synthetic and one from the FFHQ dataset -- and were asked to choose the real one from the two.
    } 
    \label{tab:real_fake_user_study}
\end{table}
\begin{table}
    \setlength{\textwidth}{0pt}
    \newcolumntype{?}{!{\vrule width 1pt}}
    \centering
    \small
	\begin{tabular}{l?c|c|c|c}
    \toprule
	Control & Ours & DFG
	& CONFIG
	& FFHQ
	\\
    \hline
    Pose [$\degree$] & $2.29$\tiny{$\pm1.31$} & $3.92$\tiny{$\pm2.1$} & $6.9$\tiny{$\pm4.7$} & $23.8$\tiny{$\pm14.6$}\\
    Age [yo]& $2.02$\tiny{$\pm1.38$} & N/A &  N/A & $16.95$\tiny{$\pm12.9$}\\
    Exp. & $3.68$\tiny{$\pm0.7$} & $4.07$\tiny{$\pm0.7$} & N/A\footnote{\label{footnote:config_exp_control}CONFIG uses different controls for expression illumination and hair color.} & $4.45$\tiny{$\pm0.9$}\\
    Illum. & $0.32$\tiny{$\pm0.13$} & $0.29$\tiny{$\pm0.1$} & N/A\textsuperscript{\ref{footnote:config_exp_control}} & $0.62$\tiny{$\pm0.2$}\\
    Hair color & $0.13$\tiny{$\pm0.18$} & N/A &  N/A\textsuperscript{\ref{footnote:config_exp_control}} & 0.34\tiny{$\pm0.25$} \\
    \bottomrule
    \end{tabular}
    \caption{\textbf{Control precision$\downarrow$:} 
    Comparison of average distance between input controls to resulted image attribute. 
    Last column shows the average distance between random samples in the FFHQ dataset.
    }
    \label{tab:precision}
\end{table}
\begin{table}
    \setlength{\textwidth}{0pt}
    \newcolumntype{?}{!{\vrule width 1pt}}
    \centering
    \small
	\begin{tabular}{l?c|c|c|c}
    \toprule
	ID & Ours & Ours$_{+age}$ & DFG
	& CONFIG
	\\
    \hline
    Same$\downarrow$ & $\pmb{0.68}$\tiny{$\pmb{\pm0.19}$} & $0.75$\tiny{$\pm0.2$} & $0.83$\tiny{$\pm0.3$} & $1.07$\tiny{$\pm0.29$} 
    \\
    Not same$\uparrow$ & \textbf{$\pmb{1.9}$\tiny{$\pmb{\pm0.24}$}} & $\pmb{1.9}$\tiny{$\pmb{\pm0.24}$} & $1.73$\tiny{$\pm0.24$} &  $1.63$\tiny{$\pm0.25$}
    \\
    \bottomrule
    \end{tabular}
    \caption{\textbf{Identity preservation:} 
    First row shows the mean embedding distance between generated images with the same $\mathbf{z}^{ID}$ (in Ours$_{+age}$, $\mathbf{z}^{age}$ is also changed). 
    Second row shows the mean embedding distance between randomly generated images. 
    For comparison, the mean embedding distance between 10K FFHQ images is $1.89\pm0.21$.}
    \label{tab:id_precision}
\end{table}

\vspace{-0.5cm}
\textbf{Photorealism:} Table~\ref{tab:fid} shows FID~\cite{heusel2017gans} scores of DiscoFaceGAN~\cite{deng2020disentangled}, CONFIG~\cite{KowalskiECCV2020} (image resolution 256x256) and our approach (image resolution 512x512).
The table also shows the FID of the corresponding baseline GANs: StyleGAN~\cite{karras2019style}, HoloGAN~\cite{HoloGAN2019} and StyleGAN2~\cite{Karras2019stylegan2}. 
Our FID score is calculated without the use of the truncation trick~\cite{brock2018large,karras2019style}. 
For DFG and CONFIG the FID score is taken from the corresponding papers.
Similarly to the other works, we observe a deterioration in FID when control is introduced.
However, due to the different image resolutions and data pre-processing steps, the numbers are not directly comparable. 
To make a clear image quality comparison between all three methods, we conducted two photorealism user studies, using Amazon Mechanical Turk.
In the first, users were shown 1K triplets of synthetic images, one from each method in a random order.
Users were asked to vote for the most realistic image of the three.
Each triplet was evaluated by three participants.
In the second study, users were shown 999 pairs of images.
Each pair contains one real image from the FFHQ dataset and an image generated by one of the three methods.
For each method, 333 image pairs were evaluated by three different users.
All the synthetic images in this experiment were generated using the truncation trick with $\Psi=0.7$ (Ours and DFG use the attribute-preserving truncation trick~\cite{deng2020disentangled}), and all images were resized to 256x256 resolution.
From Table~\ref{tab:real_fake_user_study} it is evident that our method achieves the highest photorealism.
Surprisingly, our method reaches a near perfect result of $47\%$ when compared to FFHQ, \ie, users were barely able to distinguish between our images and the ones from FFHQ in terms of photorealism.
We note that differences in image quality may depend on the base model that was used (HoloGAN, StyleGAN, StyleGAN2).
\begin{figure}[t]
\newcolumntype{Y}{>{\centering\arraybackslash}X}

\setlength{\tabcolsep}{0pt}
\begin{tabular}{c}
    \footnotesize
    \begin{tabularx}{0.98\linewidth}{ Y Y Y Y Y }
    Age$=\!15$yo & 
    $30$yo &
    $45$yo &
    $60$yo & 
    $75$yo
    \end{tabularx}
    \\
    \includegraphics[width=0.98\linewidth]{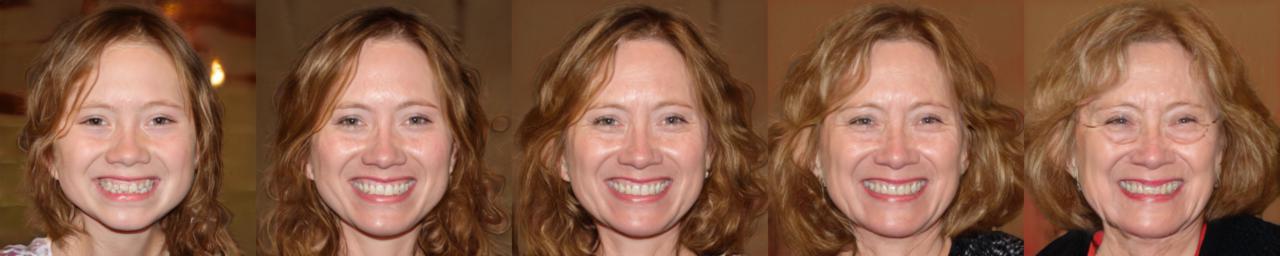} \\[-3.4pt]
    \includegraphics[width=0.98\linewidth]{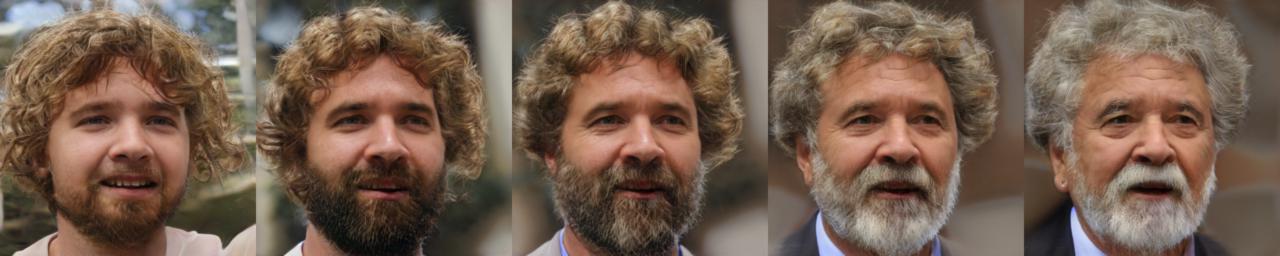} \\
\end{tabular}

\setlength{\tabcolsep}{0pt}
\begin{tabular}{c}
    \footnotesize
    \begin{tabularx}{\linewidth}{ Y Y Y Y Y }
    Yaw$=\!30\degree$ & 
    $\!15\degree$ &
    $\!0\degree$ &
    $\!-15\degree$ & 
    $\!-30\degree$
    \end{tabularx}
    \\
    \includegraphics[width=0.98\linewidth]{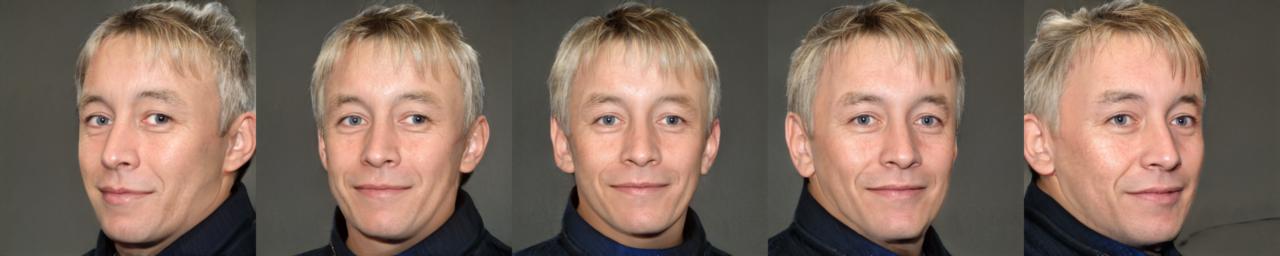} \\
    \footnotesize
    \begin{tabularx}{\linewidth}{ Y Y Y Y Y }
    Pitch$=\!20\degree$ & 
    $\!10\degree$ &
    $\!0\degree$ &
    $\!-10\degree$ & 
    $\!-20\degree$
    \end{tabularx}
    \\
    \includegraphics[width=0.98\linewidth]{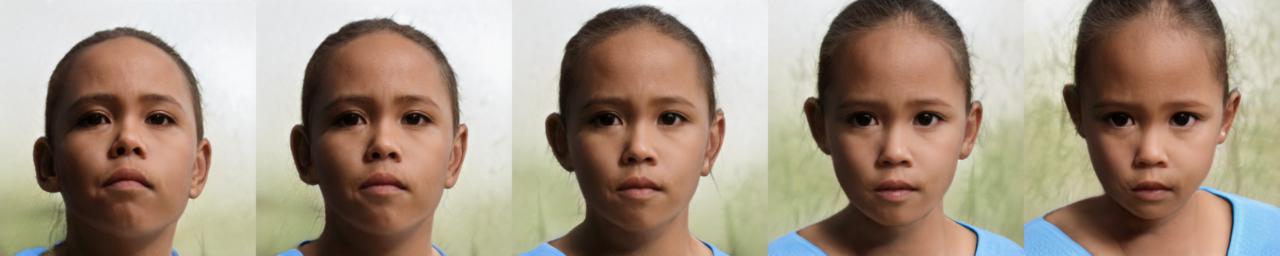} \\
\end{tabular}
\vspace{-0.3cm}
\caption{\textbf{Controlling age and pose:} Rows 1-2 show generation results using $E_{age}$. Rows 3-4 show generation results using $E_{pose}$.}
\label{fig:age_pose_control}
\end{figure}
\begin{figure}[htbp!]
\definecolor{color1}{rgb}{0.8688159, 0.8030458, 0.74361145}
\definecolor{color2}{rgb}{0.49344787, 0.4094845, 0.29451284}
\definecolor{color3}{rgb}{0.1828706, 0.14241609, 0.11705342}
\definecolor{color4}{rgb}{0.66095257, 0.52326614, 0.36687773}
\definecolor{color5}{rgb}{0.30295825, 0.28658283, 0.26644987}

\newcolumntype{Y}{>{\centering\arraybackslash}X}

\setlength{\tabcolsep}{0pt}
\begin{tabular}{c}
    \footnotesize
    \begin{tabularx}{\linewidth}{ Y Y Y Y Y }
        Illum. $1$ & 
        Illum. $2$ &
        Illum. $3$ &
        Illum. $4$ &
        Illum. $5$
    \end{tabularx}
    \\
    \includegraphics[width=0.98\linewidth]{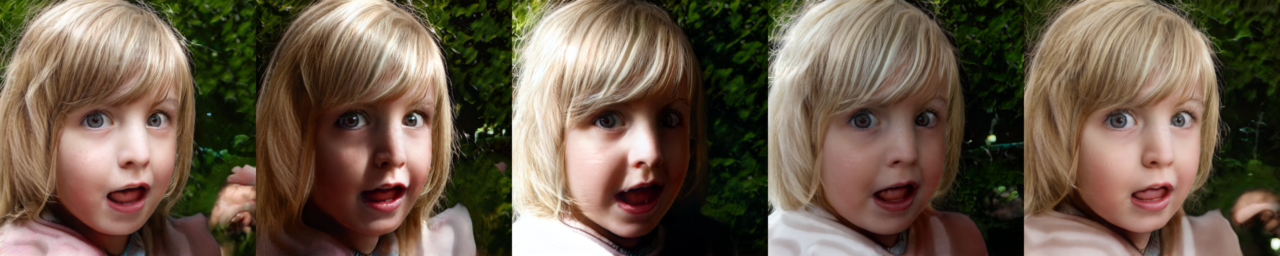} \\[-3.4pt]
    \includegraphics[width=0.98\linewidth]{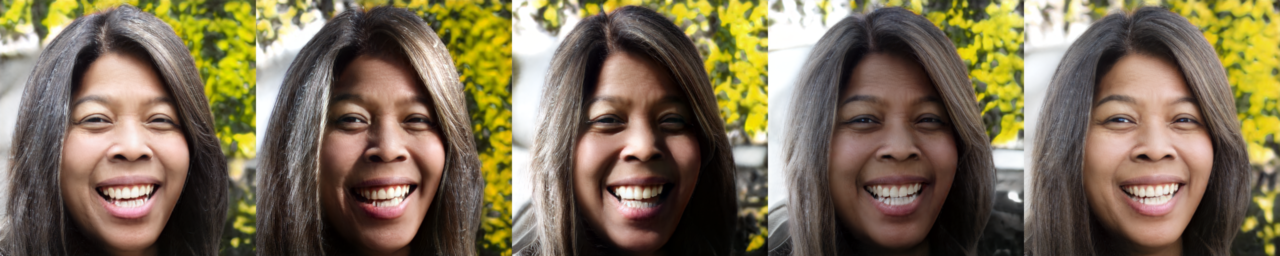} \\
\end{tabular}

\begin{tabular}{c}
    \footnotesize
    \begin{tabularx}{\linewidth}{ Y Y Y Y Y }
        Exp. $1$ & 
        Exp. $2$ &
        Exp. $3$ &
        Exp. $4$ &
        Exp. $5$
    \end{tabularx}
    \\
    \includegraphics[width=0.98\linewidth]{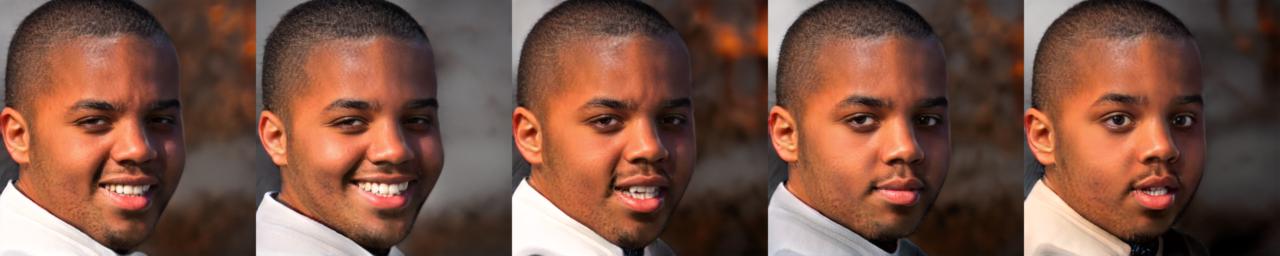} \\[-3.4pt]
    \includegraphics[width=0.98\linewidth]{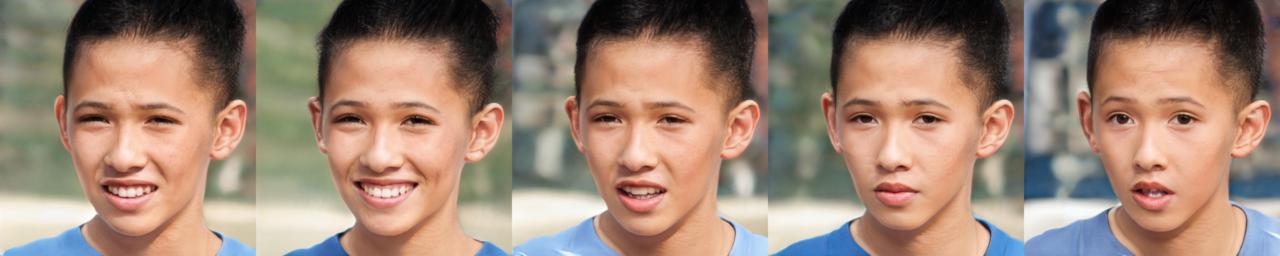} \\
\end{tabular}

\begin{tabular}{c}
    \footnotesize
    \begin{tabularx}{\linewidth}{ Y Y Y Y Y }
        \textcolor{color1}{\textbf{Color $\pmb{1}$}} &
        \textcolor{color2}{\textbf{Color $\pmb{2}$}} &
        \textcolor{color3}{\textbf{Color $\pmb{3}$}} &
        \textcolor{color4}{\textbf{Color $\pmb{4}$}} &
        \textcolor{color5}{\textbf{Color $\pmb{5}$}}
    \end{tabularx}
    \\
    \includegraphics[width=0.98\linewidth]{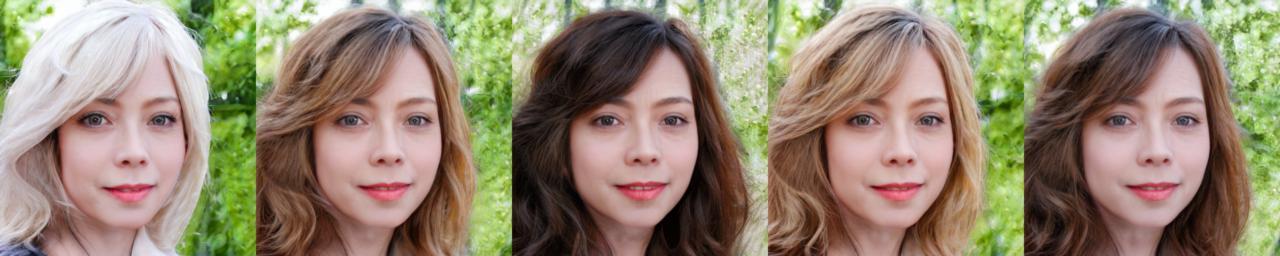} \\[-3.4pt]
    \includegraphics[width=0.98\linewidth]{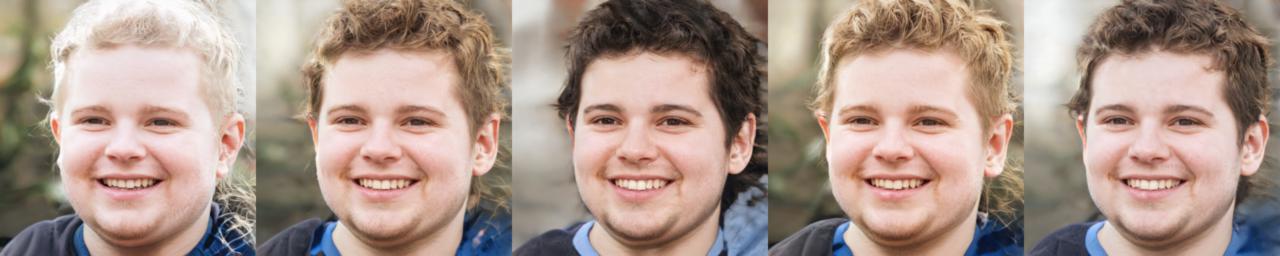} \\
\end{tabular}
\vspace{-0.3cm}
\caption{\textbf{Controlling illumination, expression and hair color:} 
Rows 1-2 show generation results using $E_{illum}$. 
Rows 3-4 show generation results using $E_{exp}$. 
Rows 5-6 show generation results using $E_{hair}$. 
Each column has the same attribute matching the control input.
}
\label{fig:controlled_illumination_expression}
\end{figure}


\textbf{Explicit control analysis:} To validate that we indeed have an explicit control over the output of our model, we perform a control precision comparison.
10K images are randomly chosen from FFHQ and their attributes are predicted to produce a pool of feasible attributes that appear in real images. 
For each attribute in the pool, $y_i^k$, we generate a corresponding image.
Then, we predict the attribute value from the generated image, $\hat{y}_i^k$, and measure the Euclidean distance between the two.
More details are provided in the supplementary material.
Table~\ref{tab:precision} shows the comparison of the control precision between the methods.
The results demonstrate that we can achieve explicit control of the attributes that is comparable or better than other methods.

\textbf{ID preservation analysis:} 
We use ArcFace~\cite{deng2019arcface} to extract embedding vectors of generated images to compare identity preservation to other methods.
This is done by generating 10K image pairs that share the ID attribute and have different pose, illumination and expression attributes.
We choose to modify these as they are common to all three methods.
To demonstrate the ability of our method to preserve the ID even at different ages, we report results for Ours$_{+age}$ where each image in a pair is generated using a different $\mathbf{z}^{age}$ vector.
The results in Table~\ref{tab:id_precision} demonstrate that our method achieves the highest identity preservation.

\textbf{Disentanglement user study:} We conducted a user study similar to the one reported in CONFIG~\cite{KowalskiECCV2020}.
For each attribute, $k$, we generate a pair of images, $\mathcal{I}_{+}, \mathcal{I}_\texttt{-} $.
The attribute for $\mathcal{I}_{+}$ is set to $y^k_{+}$ (\eg, smiling face) and the attribute for $\mathcal{I}_\texttt{-}$ is set to a semantically opposite value $y^k_\texttt{-}$ (\eg, sad face).
Users are then asked to evaluate the presence of $y^k_{+}$ in $\mathcal{I}_{+}$ and $\mathcal{I}_\texttt{-}$ on a 5-level scale.
In addition, for every pair of images the users are asked to evaluate to what extent all other attributes, apart from $k$, are preserved.
In total, 50 users have evaluated 1300 pairs of images.
Fig.~\ref{fig:user_study} clearly shows that the attributes of the generated images are perceived as disentangled.

\textbf{Qualitative evaluation:} 
Next we show editing results of generated images via the control encoders $E_k$. 
Fig.~\ref{fig:age_pose_control} shows explicit control over age and pose of faces using $E_{age}$ and $E_{pose}$. 
Interestingly, as the age is increased the model tends to generate glasses as well as more formal clothing.
Two other prominent features are graying of the hair and the addition of wrinkles.
Fig.~\ref{fig:controlled_illumination_expression} shows control over illumination and expression using $E_{illum}$ and $E_{exp}$.

\subsection{Painting generation}
\textbf{Implementation details:} We use MetFaces~\cite{karras2020training}, 1,336 images downsampled to 512x512 resolution.
In addition to the traditional StyleGAN2 and our explicit disentanglement training schemes, we use the method of non-leaking augmentation by Karras~\etal~\cite{karras2020training} for training GANs with limited data. 
We use the same $M_k$ models as in our face generation scheme with the following modifications: (1) the illumination and hair color controls are removed, (2) a control for image style is added. 
The style similarity distance $d_{style}$ is computed similarly to the style loss introduced for style transfer by Gatys~\etal~\cite{gatys2016image} where $M_{style}$ is a VGG16~\cite{DBLP:journals/corr/SimonyanZ14a} network pre-trained on ImageNet~\cite{imagenet_cvpr09}.
\begin{figure}[]
\newcolumntype{Y}{>{\centering\arraybackslash}X}

\setlength{\tabcolsep}{0pt}
\begin{tabular}{c}
    \footnotesize
    \begin{tabularx}{0.98\linewidth}{ Y Y Y Y Y }
    Age$=\!15$yo & 
    $30$yo &
    $45$yo &
    $60$yo & 
    $75$yo
    \end{tabularx}
    \\
    \includegraphics[width=0.98\linewidth]{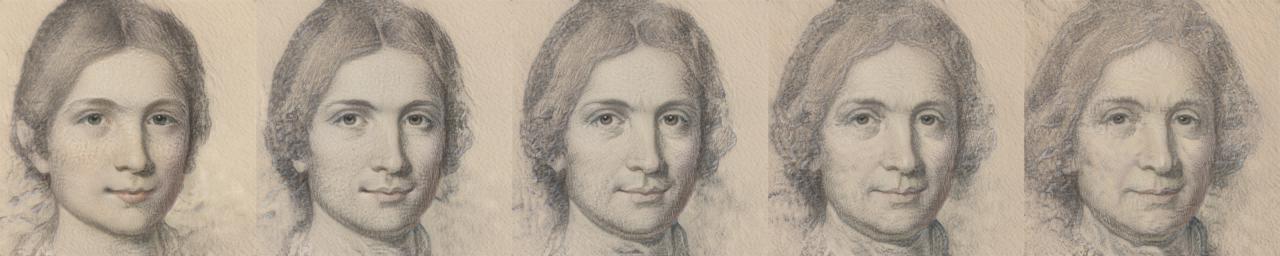} \\
\end{tabular}

\setlength{\tabcolsep}{0pt}
\begin{tabular}{c}
    \footnotesize
    \begin{tabularx}{\linewidth}{ Y Y Y Y Y }
    Yaw$=\!30\degree$ & 
    $\!15\degree$ &
    $\!0\degree$ &
    $\!-15\degree$ & 
    $\!-30\degree$
    \end{tabularx}
    \\
    \includegraphics[width=0.98\linewidth]{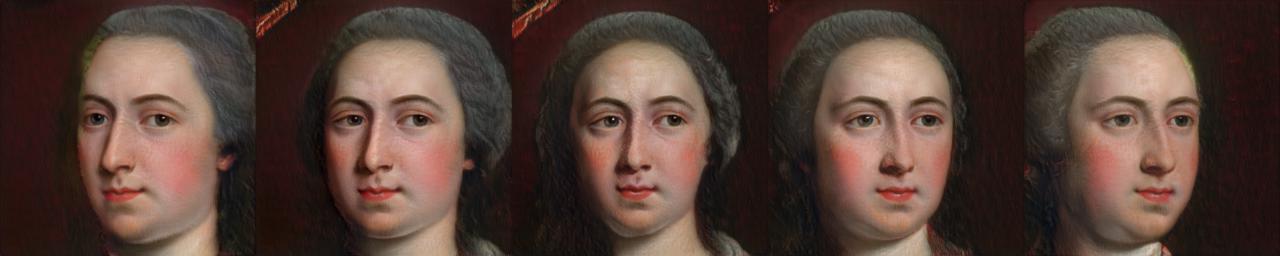} \\
\end{tabular}

\setlength{\tabcolsep}{0pt}
\begin{tabular}{c}
    \footnotesize
    \begin{tabularx}{\linewidth}{ Y Y Y Y Y }
        Exp. $1$ & 
        Exp. $2$ &
        Exp. $3$ &
        Exp. $4$ &
        Exp. $5$
    \end{tabularx}
    \\
    \includegraphics[width=0.98\linewidth]{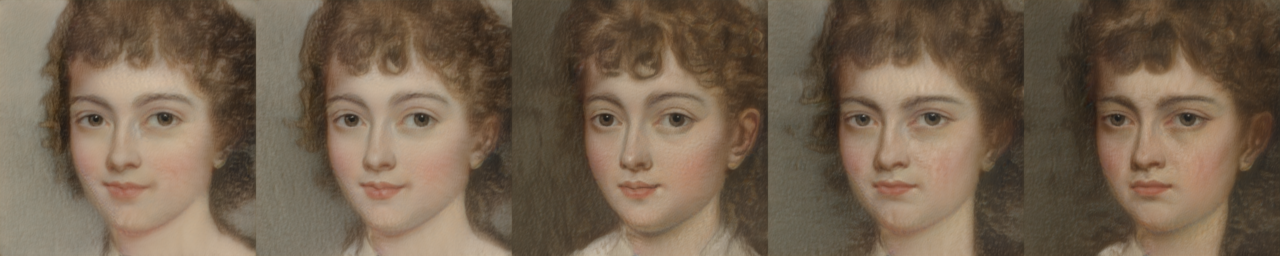} \\
\end{tabular}
\vspace{-0.2cm}
\caption{\textbf{Control of paintings:} Generation results using $E_{age}$, $E_{pose}$ and $E_{exp}$.}
\label{fig:metface_controls}
\end{figure}

\textbf{Photorealism:} 
The FID scores are $28.58$ and $26.6$ for our controlled and for the baseline models, respectively.

\begin{figure}[t]
\newcolumntype{Y}{>{\centering\arraybackslash}X}

\setlength{\tabcolsep}{0pt}
\begin{tabular}{c}
    \footnotesize
    \begin{tabularx}{0.98\linewidth}{ Y Y Y Y Y }
        $\mathbf{z}^{style}_1$ & 
        $\mathbf{z}^{style}_2$ &
        $\mathbf{z}^{style}_3$ &
        $\mathbf{z}^{style}_4$ & 
        $\mathbf{z}^{style}_5$
    \end{tabularx}
    \\
    \includegraphics[width=0.98\linewidth]{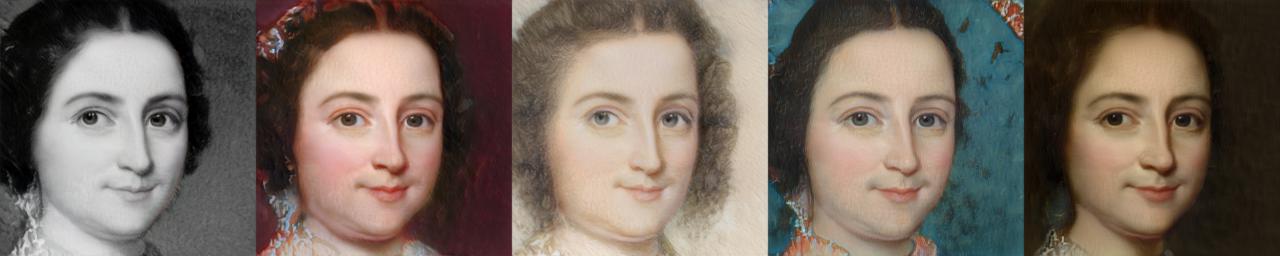} \\[-3.4pt]
    \includegraphics[width=0.98\linewidth]{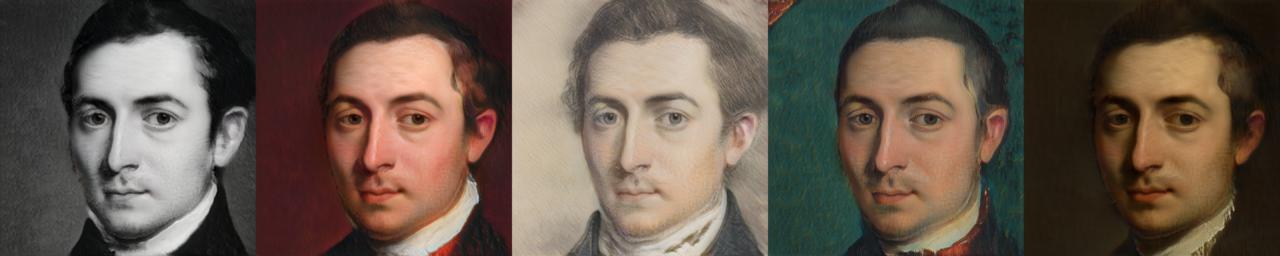} \\
\end{tabular}
\vspace{-0.2cm}
\caption{\textbf{Artistic style for paintings:} We can change the $\mathbf{z}^{style}$ latent to produce same portraits with different style.}
\label{fig:style}
\end{figure}

\textbf{Qualitative evaluation:} Fig.~\ref{fig:metface_controls} shows our control over age, pose and expression using $E_{age}$, $E_{pose}$ and $E_{exp}$. 
Note that the expression control for this task is rather limited.
We suspect this is due to the low variety of expressions in the dataset. 
The control over these attributes demonstrates that the control networks do not necessarily need to be trained on the same domain on which the GAN is being trained, and that some domain gap is tolerable.
Fig.~\ref{fig:style} shows that our method can also disentangle artistic style allowing to change the style without affecting the rest of the attributes.

\begin{table}[]
    \setlength{\textwidth}{0pt}
    \newcolumntype{?}{!{\vrule width 1pt}}
    \centering
    \small
	\begin{tabular}{l?c|c|c|c}
        \toprule
	     &  \textbf{Ours} &  E2E &  E2E-10x & NoDis
	    \\
        \hline
        \multicolumn{5}{c}{\textbf{Control precision $\downarrow$}}\\
        \hline
        Pose [$\degree$] & $\pmb{2.29}$\tiny{$\pmb{\pm1.31}$} & $10.35$\tiny{$\pm7.8$} & $4.36$\tiny{$\pm0.82$} & $5.44$\tiny{$\pm3.4$}\\
        Age [yo]& $\pmb{2.02}$\tiny{$\pmb{\pm1.38}$} & $14.63$\tiny{$\pm8.4$} & $14.38$\tiny{$\pm8.5$} & $7.11$\tiny{$\pm6.1$}\\
        Exp. & $3.68$\tiny{$\pm0.7$} & $4.41$\tiny{$\pm0.8$}  & $4.36$\tiny{$\pm0.8$} & $\pmb{2.94}$\tiny{$\pmb{\pm0.6}$}\\
        Illum. & $\pmb{0.32}$\tiny{$\pmb{\pm0.13}$} & $0.62$\tiny{$\pm0.21$} & $0.61$\tiny{$\pm0.21$} & $\pmb{0.32}$\tiny{$\pmb{\pm0.14}$}\\
        Hair c. & $\pmb{0.13}$\tiny{$\pmb{\pm0.18}$} & $0.33$\tiny{$\pm0.24$} & $0.24$\tiny{$\pm0.18$} & $0.15$\tiny{$\pm0.14$} \\
        \hline
        \multicolumn{5}{c}{\textbf{ID preservation}}\\
        \hline
        Same$\downarrow$ & 
        $\pmb{0.68}$\tiny{$\pmb{\pm0.19}$} & 
        $0.82$\tiny{$\pm0.3$} & 
        $0.97$\tiny{$\pm0.35$} & 
        $1.16$\tiny{$\pm0.34$}
        \\
        Not same$\uparrow$ & 
        $\pmb{1.9}$\tiny{$\pmb{\pm0.24}$} &
        $1.78$\tiny{$\pm0.23$} & 
        $1.79$\tiny{$\pm0.25$} & 
        $1.7$\tiny{$\pm0.26$}
        \\
        \hline
        \multicolumn{5}{c}{\textbf{FID $\downarrow$}} \\
        \hline
        FID & $5.72$ & $6.48$ & $9.1$ & $\pmb{3.32}$ \\
        \bottomrule
    \end{tabular}
    \caption{\textbf{Ablation study:} Comparison of our method vs. training end-to-end (single phase) and vs. using a non-disentangled StyleGAN2.}
    \label{tab:ablation_study}
\end{table}
\begin{figure}[t]
\newcolumntype{Y}{>{\centering\arraybackslash}X}

\definecolor{color0}{rgb}{0.7670, 0.6083, 0.5103}
\definecolor{color1}{rgb}{0.7321048 , 0.6154765 , 0.36478823}

\setlength{\tabcolsep}{0pt}
\renewcommand{\arraystretch}{0}
\begin{tabular}{c c}
    &
    \footnotesize
    \begin{tabularx}{0.98\linewidth}{Y Y Y Y Y Y }
    Initial controls &
    +Smile &
    +Age$=\!65$yo &
    +Brown hair &
    +Right Illum. &
    +Yaw$=\!0\degree$ Pitch$=\!0\degree$
    \end{tabularx}
    \\
    \rotatebox{90}{
    \begin{tabularx}{0.14\linewidth}{ Y }
        \tiny
        NoDis
    \end{tabularx}} &
    \includegraphics[width=0.98\linewidth]{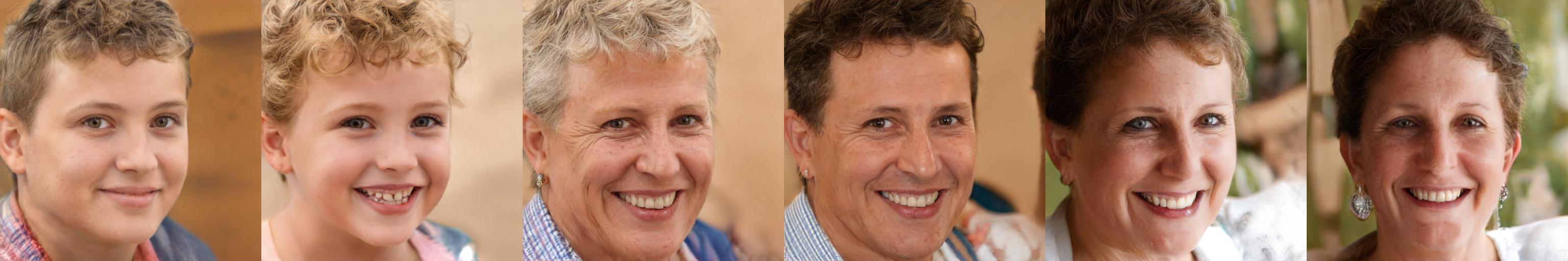} \\
    \rotatebox{90}{
    \begin{tabularx}{0.14\linewidth}{ Y }
        \tiny
        \textbf{Ours}
    \end{tabularx}} &
    \includegraphics[width=0.98\linewidth]{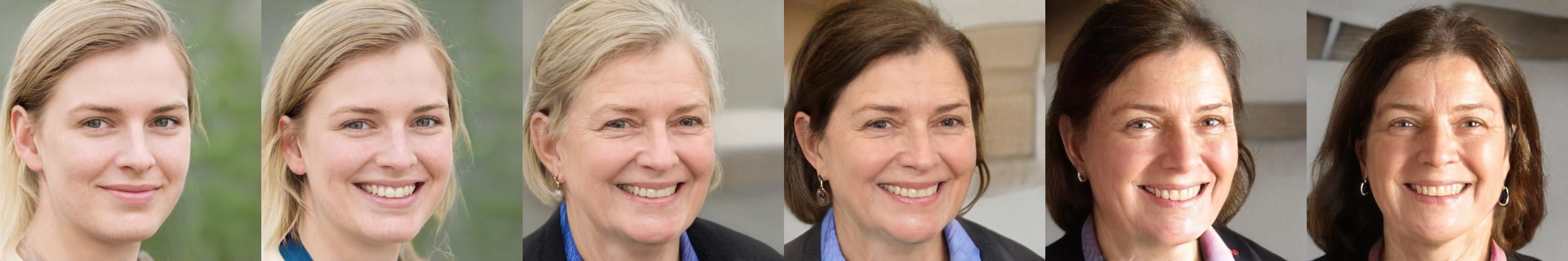} \\

\end{tabular}

\vspace{0.25cm}
\renewcommand{\arraystretch}{0.25}
\begin{tabular}{c c}
    &
    \footnotesize
    \begin{tabularx}{0.98\linewidth}{Y Y Y Y Y Y }
    ID 1 &
    ID 2 &
    ID 3 &
    ID 4 &
    ID 5 &
    ID 6
    \end{tabularx}
    \\
    \rotatebox{90}{
    \begin{tabularx}{0.14\linewidth}{ Y }
        \tiny
        NoDis
    \end{tabularx}} &
    \includegraphics[width=0.98\linewidth]{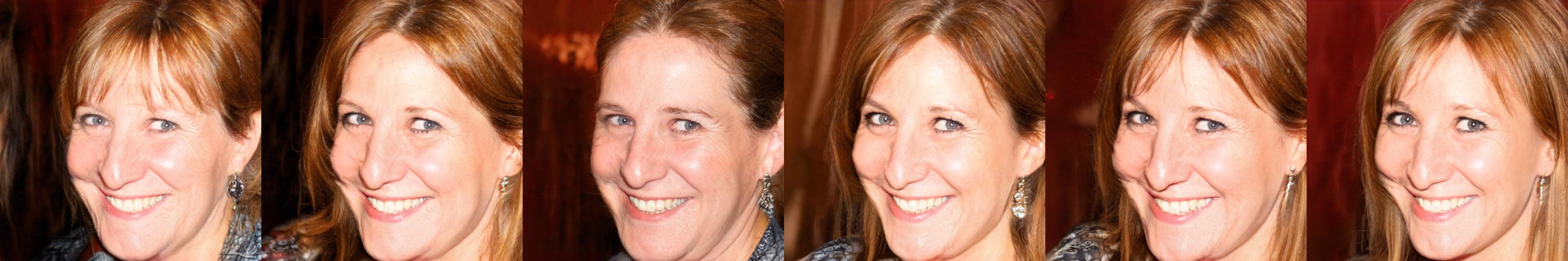} 
    
\end{tabular}

\vspace{-0.2cm}
\footnotesize
\caption{\textbf{Ours vs. NoDis:} In row 1 (NoDis) and row 2 (ours), from left to right, each column changes one control. The ID input is not changed. In row 3 (NoDis), each column has a different ID input and same control inputs.  
}
\label{fig:r4_vs_ours}
\end{figure}
\subsection{Ablation study}
\label{sec:ablation study}
In this section we explore two alternative approaches to our framework.
(1) Training the GAN end-to-end in a single training phase.
In every iteration, the inputs to the model are control attribute values, ($y^k$), rather than latent vectors.
We use the pre-trained models (the same ones as in our two-phase approach) to penalize for disagreement between the attribute values, predicted for each generated image, and the input controls (attribute matching loss).
For a fair comparison to our approach, we avoid the harder task of mapping an ID embedding to an image, by maintaining the ID contrastive terms as in Sec.~\ref{sec:loss}.
We use two configurations of matching loss coefficients where for the first model (E2E) the coefficients are 10 times smaller in magnitude than for the second one (E2E-10x).
(2) Instead of training a disentangled GAN in Phase 1, we train attribute encoders for a pre-trained StyleGAN2 (NoDis).
Since StyleGAN2's $\mathcal{W}$ space is not
divided into disentangled sub-spaces, 
we train a single encoder mapping all inputs (together) to $\mathcal{W}$. 
Further implementation details of alternatives 1 and 2 are provided in the supplementary.

In Table~\ref{tab:ablation_study} we compare our two-phased approach to both alternatives using the control precision, ID preservation and FID metrics.
As expected, the E2E-10x model achieves better control precision than the E2E model at the expense of reduced photorealism (FID score) and ID preservation.
Nonetheless, at both ends of the spectrum the results are inferior to those achieved by our two-phased model.
We present qualitative comparisons in the supplementary.
Table~\ref{tab:ablation_study} indicates that NoDis does not preserve ID.
This is backed up by the first row of Fig.~\ref{fig:r4_vs_ours}.
In the third row of Fig.~\ref{fig:r4_vs_ours} we show images generated for different ID vectors but with the same set of controls.
The mild variation in perceived ID demonstrates that the entanglement limits the possible IDs, given a set of controls.
Moreover, for NoDis the control precision is inferior except for the expression.
We hypothesize that in order to reach a desired control, the model partially ``adjusts the ID''.
This is most prominent for expression where the geometry of the face changes.
Thus with limited ID preservation, it is ``easier'' to achieve a desired expression.

\subsection{Disentangled projection of real images}
We leverage the explicit control of our model for real image editing.
To this end, we use latent space optimization to find a latent vector that corresponds to an input image.
By na\text{\"i}vely following the projection method described in StyleGAN2 (Appendix D), the reconstructed image visually looks different from the input image.
A remedy to this phenomenon proposed in~\cite{Abdal_2019_ICCV} is to project the image to an  extended latent space, $w^+$, such that each resolution level has its own latent vector.
We find this approach is indeed useful for accurate reconstruction.
However, when we modified the different sub-vectors, we observed a strong deterioration in the image quality and a change in other unmodified attributes.
In absence of explicit constraints on the feasible solutions' space, two different issues arise:
(1) part of the sub-vectors end-up encoding a semantically different information from the one they were intended for, \eg, the pose latent vector may encode some information of the ID or the expression, and
(2) the reconstructed latent vector may not lie in the semantically meaningful manifold.
A similar phenomenon was reported in Zhu~\etal~\cite{zhu2020indomain}.
As a mitigation to the above, we introduce two changes.
First, rather than extending the entire $\mathcal{W}$ space, we only extend the $\mathcal{W}^{ID}$ and $\mathcal{W}^{other}$ sub-spaces.
Second, we constrain the remaining sub-vectors to reside on approximated linear subspaces of their corresponding manifolds. 
We achieve this using the following approach: we perform PCA for each latent subspace of 10K randomly sampled sub-vectors $\mathbf{w}$, where the number of components are selected so as to preserve 50\% of the variance.
During the optimization process, we project the latent sub-vectors to the truncated PCA spaces and re-project them back to the corresponding spaces.
Once we find the corresponding latent vector, we can edit the image by modifying attribute $k$ latent sub-vector, using $E_k$.
We provide an ablation study of the proposed changes in the supplementary material.

In Fig.~\ref{fig:projection_ori_gamma} we show real images, their projections and the result of editing their attributes.
While the projected image does not achieve a perfect reconstruction, the disentanglment of the latent space is preserved, allowing for an explicit control of the desired attributes without affecting others.
In the second row of Fig.~\ref{fig:projection_ori_gamma} we can see that the GAN can accurately model the shadows on the face's curvature and skin folds as well as model the reflection of the light source in the person's eyes.
This implies the GAN learns a latent 3D representation of the faces.


\begin{figure}[]
\newcolumntype{Y}{>{\centering\arraybackslash}X}

\setlength{\tabcolsep}{0pt}
\begin{tabular}{c}
    \footnotesize
    \begin{tabularx}{\linewidth}{ Y Y Y Y Y }
    Input~\cite{FFHQ_12} &
    Projected &
    Yaw$=\!30\degree$ Pitch$=\!0\degree$ & 
    Yaw$=\!25\degree$ Pitch$=\!-15\degree$ & 
    Yaw$=\!-25\degree$ Pitch$=\!15\degree$ 
    \end{tabularx}
    \\
    \includegraphics[width=0.98\linewidth]{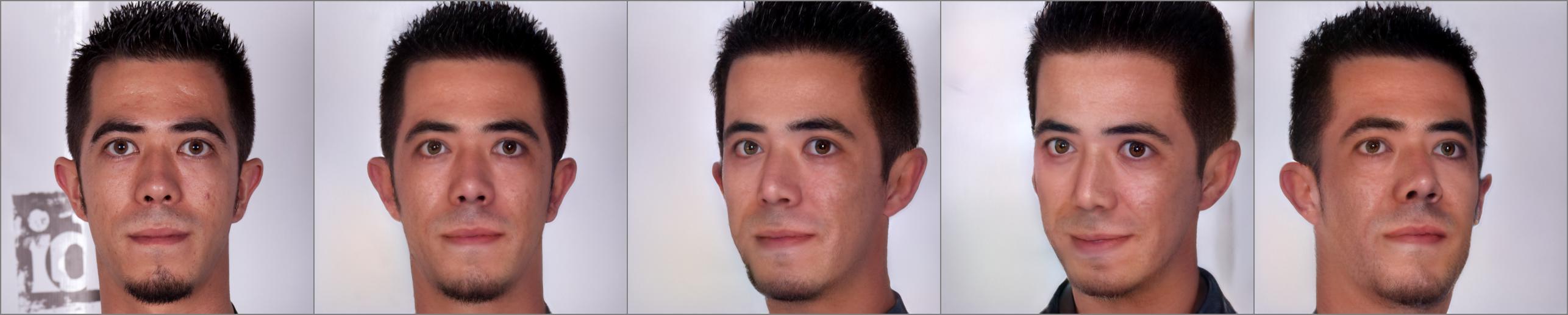} \\
    \footnotesize
    \begin{tabularx}{\linewidth}{ Y Y Y Y Y }
    Input~\cite{FFHQ_247} &
    Projected &
    Right &
    Front & 
    Left 
    \end{tabularx}
    \\
    \includegraphics[width=0.98\linewidth]{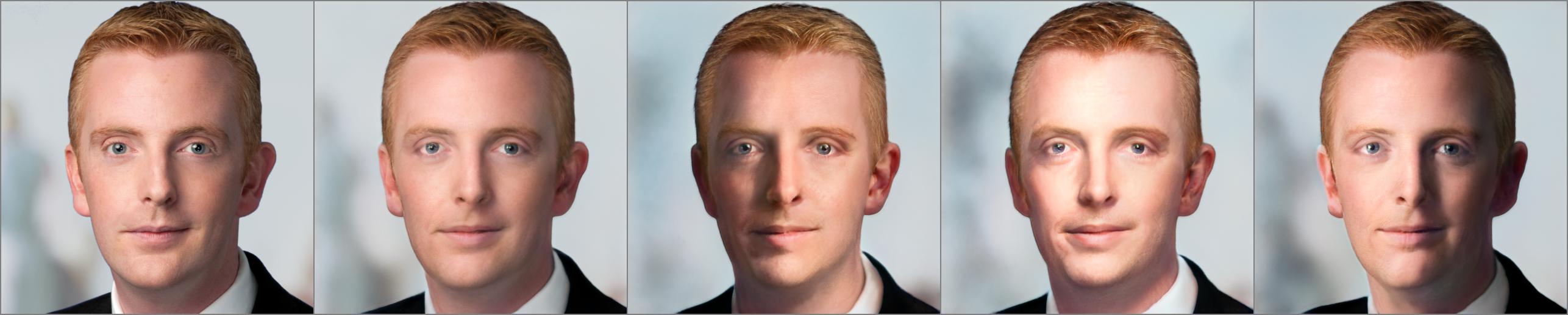} \\
    \footnotesize
    \begin{tabularx}{\linewidth}{ Y Y Y Y Y }
    Input~\cite{FFHQ_200} &
    Projected &
    Age$=\!15$yo &
    $45$yo & 
    $70$yo
    \end{tabularx}
    \\
    \includegraphics[width=0.98\linewidth]{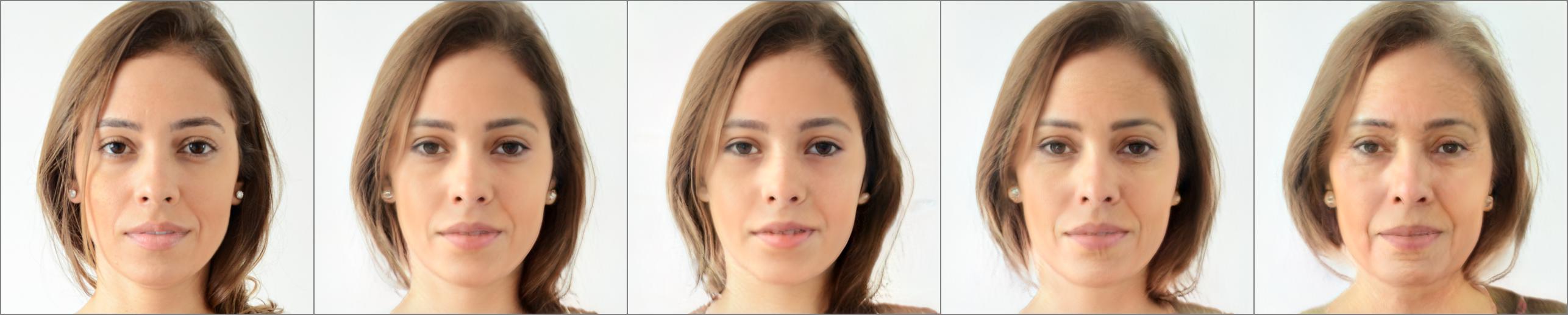} \\
    \footnotesize
    \begin{tabularx}{\linewidth}{ Y Y Y Y Y }
    Input~\cite{FFHQ_114} &
    Projected &
    Exp. 1 &
    Exp. 2 & 
    Exp. 3
    \end{tabularx}
    \\
    \includegraphics[width=0.98\linewidth]{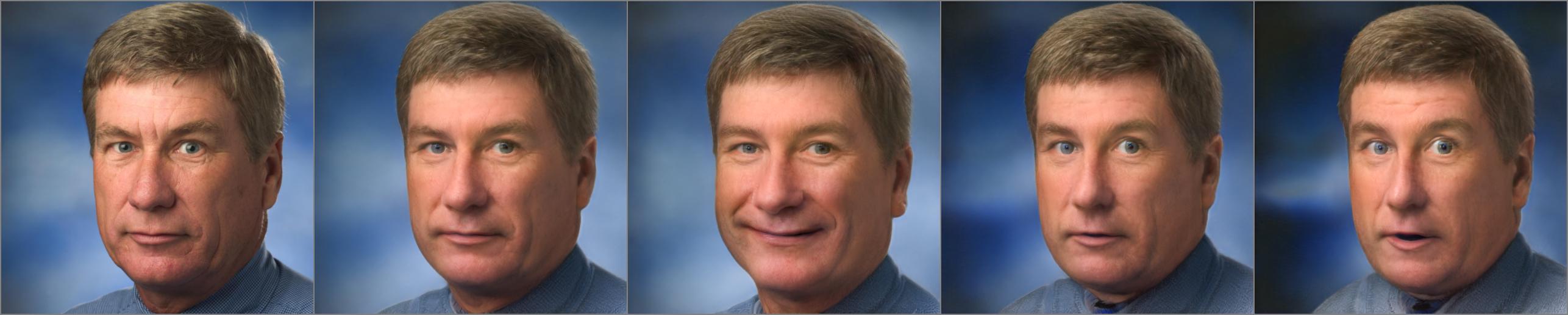} \\
\end{tabular}
\vspace{-0.3cm}
\caption{\textbf{Disentangled Projection:} 
The two leftmost columns refer to the input and projected images, respectively.
The remaining columns demonstrate editing results of pose, illumination, age and expression.
}
\label{fig:projection_ori_gamma}
\end{figure}


\section{Conclusions}
\label{sec:conclusions}
We proposed a novel framework for training GANs in a disentangled manner, that allows explicit control over generation attributes. 
For a variety of attributes, a predictor of that attribute is enough to achieve explicit control over it.
Our method extends the applicability of explicitly controllable GANs to additional domains other than human faces.
The GAN is complemented by a real image projection method that projects images to a disentangled latent space, maintaining explicit control. 
We believe this work opens up a path for improving the ability to control general-purpose GAN generation.
Additional details can be found at \url{alonshoshan10.github.io/gan_control/}.

\newpage

{\small
\bibliographystyle{ieee_fullname}
\bibliography{egbib}

\begin{thebibliography}{10}\itemsep=-1pt

\bibitem{FFHQ_12}
The original image is at \url{http://www.flickr.com/photos/quakecon/3923570806}
  and is licensed under: \url{http://www.creativecommons.org/licenses/by/2.0}.

\bibitem{FFHQ_247}
The original image is at
  \url{http://www.flickr.com/photos/dereknolan/5309847731} and is licensed
  under: \url{http://www.creativecommons.org/licenses/by/2.0}.

\bibitem{FFHQ_200}
The original image is at
  \url{http://www.flickr.com/photos/67548743@N02/6854926480} and is licensed
  under: \url{http://www.creativecommons.org/licenses/by/2.0}.

\bibitem{FFHQ_114}
The original image is at
  \url{http://www.flickr.com/photos/ugacommunications/6005899336} and is
  licensed under: \url{http://www.creativecommons.org/licenses/by-nc/2.0}.

\bibitem{Abdal_2019_ICCV}
Rameen Abdal, Yipeng Qin, and Peter Wonka.
\newblock Image2stylegan: How to {E}mbed {I}mages {I}nto the {S}tyle{GAN}
  {L}atent {S}pace?
\newblock In {\em Proceedings of the IEEE/CVF International Conference on
  Computer Vision (ICCV)}, October 2019.

\bibitem{abdal2020styleflow}
Rameen Abdal, Peihao Zhu, Niloy Mitra, and Peter Wonka.
\newblock Styleflow: Attribute-conditioned exploration of stylegan-generated
  images using conditional continuous normalizing flows, 2020.

\bibitem{balakrishnan2020towards}
Guha Balakrishnan, Yuanjun Xiong, Wei Xia, and Pietro Perona.
\newblock Towards {C}ausal {B}enchmarking of {B}ias in {F}ace {A}nalysis
  {A}lgorithms.
\newblock 2020.

\bibitem{Bao0WLH18}
Jianmin Bao, Dong Chen, Fang Wen, Houqiang Li, and Gang Hua.
\newblock Towards open-set identity preserving face synthesis.
\newblock In {\em 2018 {IEEE} Conference on Computer Vision and Pattern
  Recognition, {CVPR} 2018, Salt Lake City, UT, USA, June 18-22, 2018}, pages
  6713--6722. {IEEE} Computer Society, 2018.

\bibitem{blanz1999morphable}
Volker Blanz and Thomas Vetter.
\newblock {A Morphable Model for the Synthesis of 3D Faces}.
\newblock In {\em Proceedings of the 26th Annual Conference on Computer
  Graphics and Interactive Techniques}, pages 187--194, 1999.

\bibitem{brock2018large}
Andrew Brock, Jeff Donahue, and Karen Simonyan.
\newblock Large {S}cale {GAN} {T}raining for {H}igh {F}idelity {N}atural
  {I}mage {S}ynthesis.
\newblock In {\em International Conference on Learning Representations}, 2019.

\bibitem{choi2018stargan}
Yunjey Choi, Minje Choi, Munyoung Kim, Jung-Woo Ha, Sunghun Kim, and Jaegul
  Choo.
\newblock Star{GAN}: {U}nified {G}enerative {A}dversarial {N}etworks for
  {M}ulti-{D}omain {I}mage-to-{I}mage {T}ranslation.
\newblock In {\em Proceedings of the IEEE Conference on Computer Vision and
  Pattern Recognition {CVPR}}, 2018.

\bibitem{choi2020starganv2}
Yunjey Choi, Youngjung Uh, Jaejun Yoo, and Jung-Woo Ha.
\newblock {StarGAN v2: Diverse {I}mage {S}ynthesis for {M}ultiple {D}omains}.
\newblock In {\em Proceedings of the IEEE Conference on Computer Vision and
  Pattern Recognition {CVPR}}, 2020.

\bibitem{imagenet_cvpr09}
J. Deng, W. Dong, R. Socher, L.-J. Li, K. Li, and L. Fei-Fei.
\newblock {ImageNet: A Large-Scale Hierarchical Image Database}.
\newblock In {\em IEEE/CVF Conference on Computer Vision and Pattern
  Recognition (CVPR)}, 2009.

\bibitem{deng2019arcface}
Jiankang Deng, Jia Guo, Niannan Xue, and Stefanos Zafeiriou.
\newblock Arc{F}ace: {A}dditive {A}ngular {M}argin {L}oss for {D}eep {F}ace
  {R}ecognition.
\newblock In {\em Proceedings of the IEEE Conference on Computer Vision and
  Pattern Recognition}, pages 4690--4699, 2019.

\bibitem{deng2020disentangled}
Yu Deng, Jiaolong Yang, Dong Chen, Fang Wen, and Xin Tong.
\newblock Disentangled and {C}ontrollable {F}ace {I}mage {G}eneration via 3{D}
  {I}mitative-{C}ontrastive {L}earning.
\newblock In {\em IEEE Computer Vision and Pattern Recognition}, 2020.

\bibitem{deng2019accurate}
Yu Deng, Jiaolong Yang, Sicheng Xu, Dong Chen, Yunde Jia, and Xin Tong.
\newblock Accurate 3{D} {F}ace {R}econstruction with {W}eakly-{S}upervised
  {L}earning: from {S}ingle {I}mage to {I}mage {S}et.
\newblock In {\em IEEE Computer Vision and Pattern Recognition Workshops},
  2019.

\bibitem{ding2021ccgan}
Xin Ding, Yongwei Wang, Zuheng Xu, William~J Welch, and Z.~Jane Wang.
\newblock Cc{\{}gan{\}}: Continuous conditional generative adversarial networks
  for image generation.
\newblock In {\em International Conference on Learning Representations}, 2021.

\bibitem{donahue2018sdgan}
Chris Donahue, Zachary~C Lipton, Akshay Balsubramani, and Julian McAuley.
\newblock Semantically {D}ecomposing the {L}atent {S}paces of {G}enerative
  {A}dversarial {N}etworks.
\newblock In {\em ICLR}, 2018.

\bibitem{egger20203d}
Bernhard Egger, William~AP Smith, Ayush Tewari, Stefanie Wuhrer, Michael
  Zollhoefer, Thabo Beeler, Florian Bernard, Timo Bolkart, Adam Kortylewski,
  Sami Romdhani, et~al.
\newblock 3{D} {M}orphable {F}ace {M}odels—{P}ast, {P}resent, and {F}uture.
\newblock {\em ACM Transactions on Graphics (TOG)}, 39(5):1--38, 2020.

\bibitem{gatys2016image}
Leon~A Gatys, Alexander~S Ecker, and Matthias Bethge.
\newblock Image {S}tyle {T}ransfer {U}sing {C}onvolutional {N}eural {N}etworks.
\newblock In {\em CVPR}, 2016.

\bibitem{goodfellow2014generative}
Ian Goodfellow, Jean Pouget-Abadie, Mehdi Mirza, Bing Xu, David Warde-Farley,
  Sherjil Ozair, Aaron Courville, and Yoshua Bengio.
\newblock Generative {A}dversarial {N}ets.
\newblock In {\em Advances in {N}eural {I}nformation {P}rocessing {S}ystems},
  pages 2672--2680, 2014.

\bibitem{harkonen2020ganspace}
Erik H{\"a}rk{\"o}nen, Aaron Hertzmann, Jaakko Lehtinen, and Sylvain Paris.
\newblock Ganspace: {D}iscovering {I}nterpretable {GAN} {C}ontrols.
\newblock {\em arXiv preprint arXiv:2004.02546}, 2020.

\bibitem{AttGan}
Zhenliang He, Wangmeng Zuo, Meina Kan, Shiguang Shan, and Xilin Chen.
\newblock Attgan: Facial attribute editing by only changing what you want.
\newblock {\em {IEEE} Trans. Image Process.}, 28(11):5464--5478, 2019.

\bibitem{heusel2017gans}
Martin Heusel, Hubert Ramsauer, Thomas Unterthiner, Bernhard Nessler, and Sepp
  Hochreiter.
\newblock Gans {T}rained by a {T}wo {T}ime-{S}cale {U}pdate {R}ule {C}onverge
  to a {L}ocal {N}ash {E}quilibrium.
\newblock In {\em NIPS}, 2017.

\bibitem{huang2018munit}
Xun Huang, Ming-Yu Liu, Serge Belongie, and Jan Kautz.
\newblock Multimodal {U}nsupervised {I}mage-to-{I}mage {T}ranslation.
\newblock In {\em ECCV}, 2018.

\bibitem{gansteerability}
Ali Jahanian, Lucy Chai, and Phillip Isola.
\newblock On the "{S}teerability" of {G}enerative {A}dversarial {N}etworks.
\newblock In {\em International Conference on Learning Representations}, 2020.

\bibitem{karras2017progressive}
Tero Karras, Timo Aila, Samuli Laine, and Jaakko Lehtinen.
\newblock Progressive {G}rowing of {GAN}s for {I}improved {Q}uality,
  {S}tability, and {V}ariation.
\newblock {\em arXiv preprint arXiv:1710.10196}, 2017.

\bibitem{karras2020training}
Tero Karras, Miika Aittala, Janne Hellsten, Samuli Laine, Jaakko Lehtinen, and
  Timo Aila.
\newblock Training generative adversarial networks with limited data.
\newblock {\em arXiv preprint arXiv:2006.06676}, 2020.

\bibitem{karras2019style}
Tero Karras, Samuli Laine, and Timo Aila.
\newblock A {S}tyle-{B}ased {G}enerator {A}rchitecture for {G}enerative
  {A}dversarial {N}etworks.
\newblock In {\em Proceedings of the IEEE Conference on Computer Vision and
  Pattern Recognition {CVPR}}, pages 4401--4410, 2019.

\bibitem{Karras2019stylegan2}
Tero Karras, Samuli Laine, Miika Aittala, Janne Hellsten, Jaakko Lehtinen, and
  Timo Aila.
\newblock Analyzing and {I}mproving the {I}mage {Q}uality of {StyleGAN}.
\newblock {\em CoRR}, abs/1912.04958, 2019.

\bibitem{KowalskiECCV2020}
Marek Kowalski, Stephan~J. Garbin, Virginia Estellers, Tadas Baltrušaitis,
  Matthew Johnson, and Jamie Shotton.
\newblock {CONFIG: Controllable Neural Face Image Generation}.
\newblock In {\em European Conference on Computer Vision (ECCV)}, 2020.

\bibitem{li2017learning}
Tianye Li, Timo Bolkart, Michael~J Black, Hao Li, and Javier Romero.
\newblock {Learning a Model of Facial Shape and Expression from 4D Scans.}
\newblock {\em ACM Trans. Graph.}, 36(6):194--1, 2017.

\bibitem{cgan}
Mehdi Mirza and Simon Osindero.
\newblock Conditional generative adversarial nets.
\newblock {\em CoRR}, 2014.

\bibitem{Miyato2018cGANsWP}
Takeru Miyato and Masanori Koyama.
\newblock cgans with projection discriminator.
\newblock {\em ArXiv}, abs/1802.05637, 2018.

\bibitem{HoloGAN2019}
Thu Nguyen-Phuoc, Chuan Li, Lucas Theis, Christian Richardt, and Yong-Liang
  Yang.
\newblock { HoloGAN: Unsupervised Learning of 3D Representations From Natural
  Images}.
\newblock In {\em The IEEE International Conference on Computer Vision (ICCV)},
  Nov 2019.

\bibitem{Odena2017ConditionalIS}
Augustus Odena, Christopher Olah, and Jonathon Shlens.
\newblock Conditional image synthesis with auxiliary classifier gans.
\newblock In {\em ICML}, 2017.

\bibitem{nizan2020council}
Ayellet~Tal Ori~Nizan.
\newblock {Breaking the Cycle - Colleagues are All You Need}.
\newblock In {\em {IEEE Conference on Computer Vision and Pattern Recognition
  {CVPR}}}, 2020.

\bibitem{pan2021do}
Xingang Pan, Bo Dai, Ziwei Liu, Chen~Change Loy, and Ping Luo.
\newblock Do 2d {\{}gan{\}}s know 3d shape? unsupervised 3d shape
  reconstruction from 2d image {\{}gan{\}}s.
\newblock In {\em International Conference on Learning Representations}, 2021.

\bibitem{park2019SPADE}
Taesung Park, Ming-Yu Liu, Ting-Chun Wang, and Jun-Yan Zhu.
\newblock {Semantic Image Synthesis with Spatially-Adaptive Normalization}.
\newblock In {\em Proceedings of the IEEE Conference on Computer Vision and
  Pattern Recognition {CVPR}}, 2019.

\bibitem{Ramamoorthi2001AnER}
R. Ramamoorthi and P. Hanrahan.
\newblock An {E}fficient {R}epresentation for {I}rradiance {E}nvironment
  {M}aps.
\newblock In {\em SIGGRAPH '01}, 2001.

\bibitem{Rothe-ICCVW-2015}
Rasmus Rothe, Radu Timofte, and Luc~Van Gool.
\newblock Dex: Deep {E}xpectation of {A}pparent {A}ge from a {S}ingle {I}mage.
\newblock In {\em IEEE International Conference on Computer Vision Workshops
  (ICCVW)}, December 2015.

\bibitem{Ruiz_2018_CVPR_Workshops}
Nataniel Ruiz, Eunji Chong, and James~M. Rehg.
\newblock {Fine-Grained Head Pose Estimation Without Keypoints}.
\newblock In {\em The IEEE Conference on Computer Vision and Pattern
  Recognition (CVPR) Workshops}, June 2018.

\bibitem{sanyal2019learning}
Soubhik Sanyal, Timo Bolkart, Haiwen Feng, and Michael~J Black.
\newblock {Learning to Regress 3D Face Shape and Expression from an Image
  without 3D Supervision}.
\newblock In {\em Proceedings of the IEEE Conference on Computer Vision and
  Pattern Recognition}, pages 7763--7772, 2019.

\bibitem{Shen_2020_CVPR}
Yujun Shen, Jinjin Gu, Xiaoou Tang, and Bolei Zhou.
\newblock {Interpreting the Latent Space of GANs for Semantic Face Editing}.
\newblock In {\em IEEE/CVF Conference on Computer Vision and Pattern
  Recognition (CVPR)}, June 2020.

\bibitem{DBLP:conf/cvpr/Shen0YWT18}
Yujun Shen, Ping Luo, Junjie Yan, Xiaogang Wang, and Xiaoou Tang.
\newblock {FaceID-GAN: Learning a Symmetry Three-Player GAN for
  Identity-Preserving Face Synthesis}.
\newblock In {\em IEEE Conference on Computer Vision and Pattern Recognition,
  CVPR 2018}, 2018.

\bibitem{FaceFeat-GAN}
Yujun Shen, Bolei Zhou, Ping Luo, and Xiaoou Tang.
\newblock Facefeat-gan: a two-stage approach for identity-preserving face
  synthesis.
\newblock {\em CoRR}, abs/1812.01288, 2018.

\bibitem{DBLP:journals/corr/SimonyanZ14a}
Karen Simonyan and Andrew Zisserman.
\newblock Very {D}eep {C}onvolutional {N}etworks for {L}arge-{S}cale {I}mage
  {R}ecognition.
\newblock In {\em 3rd International Conference on Learning Representations,
  {ICLR}}, 2015.

\bibitem{SMW20}
Henrique Siqueira, Sven Magg, and Stefan Wermter.
\newblock {Efficient Facial Feature Learning with Wide Ensemble-based
  Convolutional Neural Networks}, Feb 2020.

\bibitem{spingarn2021gan}
Nurit Spingarn, Ron Banner, and Tomer Michaeli.
\newblock {\{}GAN{\}} ''steerability'' without optimization.
\newblock In {\em International Conference on Learning Representations}, 2021.

\bibitem{Tewari_2020_CVPR}
Ayush Tewari, Mohamed Elgharib, Gaurav Bharaj, Florian Bernard, Hans-Peter
  Seidel, Patrick Perez, Michael Zollhofer, and Christian Theobalt.
\newblock {StyleRig: Rigging StyleGAN for 3D Control Over Portrait Images}.
\newblock In {\em Proceedings of the IEEE/CVF Conference on Computer Vision and
  Pattern Recognition (CVPR)}, June 2020.

\bibitem{tewari2020pie}
Ayush Tewari, Mohamed Elgharib, Mallikarjun BR, Florian Bernard, Hans-Peter
  Seidel, Patrick P{\'e}rez, Michael Z{\"o}llhofer, and Christian Theobalt.
\newblock Pie: Portrait image embedding for semantic control.
\newblock volume~39, December 2020.

\bibitem{wang2018pix2pixHD}
Ting-Chun Wang, Ming-Yu Liu, Jun-Yan Zhu, Andrew Tao, Jan Kautz, and Bryan
  Catanzaro.
\newblock {High-Resolution Image Synthesis and Semantic Manipulation with
  Conditional GANs}.
\newblock In {\em Proceedings of the IEEE Conference on Computer Vision and
  Pattern Recognition {CVPR}}, 2018.

\bibitem{yang2019semantic}
Ceyuan Yang, Yujun Shen, and Bolei Zhou.
\newblock {Semantic Hierarchy Emerges in Deep Generative Representations for
  Scene Synthesis}.
\newblock {\em arXiv preprint arXiv:1911.09267}, 2019.

\bibitem{yao2020high}
Xu Yao, Gilles Puy, Alasdair Newson, Yann Gousseau, and Pierre Hellier.
\newblock High resolution face age editing.
\newblock {\em arXiv preprint arXiv:2005.04410}, 2020.

\bibitem{zhao2017pspnet}
Hengshuang Zhao, Jianping Shi, Xiaojuan Qi, Xiaogang Wang, and Jiaya Jia.
\newblock {Pyramid Scene Parsing Network}.
\newblock In {\em CVPR}, 2017.

\bibitem{zhu2020indomain}
Jiapeng Zhu, Yujun Shen, Deli Zhao, and Bolei Zhou.
\newblock {In-domain GAN Inversion for Real Image Editing}.
\newblock In {\em Proceedings of European Conference on Computer Vision
  (ECCV)}, 2020.

\bibitem{zhu2016generative}
Jun-Yan Zhu, Philipp Kr{\"a}henb{\"u}hl, Eli Shechtman, and Alexei~A Efros.
\newblock Generative visual manipulation on the natural image manifold.
\newblock In {\em European conference on computer vision}, pages 597--613.
  Springer, 2016.

\bibitem{CycleGAN2017}
Jun-Yan Zhu, Taesung Park, Phillip Isola, and Alexei~A Efros.
\newblock Unpaired {I}mage-to-{I}mage {T}ranslation using {C}ycle-{C}onsistent
  {A}dversarial {N}etworks.
\newblock In {\em IEEE International Conference on Computer Vision {ICCV}},
  2017.

\bibitem{zhuang2021enjoy}
Peiye Zhuang, Oluwasanmi~O Koyejo, and Alex Schwing.
\newblock Enjoy your editing: Controllable {\{}gan{\}}s for image editing via
  latent space navigation.
\newblock In {\em International Conference on Learning Representations}, 2021.

\end{thebibliography}
}

\end{document}